\documentclass{article}
\usepackage{iclr2027_conference,times}

\usepackage{amsmath,amsfonts,bm}

\def\eqref#1{equation~\ref{#1}}

\def\1{\bm{1}}

\DeclareMathAlphabet{\mathsfit}{\encodingdefault}{\sfdefault}{m}{sl}
\SetMathAlphabet{\mathsfit}{bold}{\encodingdefault}{\sfdefault}{bx}{n}

\usepackage{microtype}
\usepackage{graphicx}
\usepackage{booktabs} 
\usepackage{makecell}
\usepackage{multirow}
\usepackage{algorithm}  
\usepackage{algorithmicx}  
\usepackage{algpseudocode} 
\usepackage{xcolor}
\usepackage{colortbl}
\usepackage{enumitem}
\usepackage{pifont}

\usepackage{xcolor}
\definecolor{linkpink}{HTML}{B24C8F}

\usepackage{hyperref}
\hypersetup{
    colorlinks=true,
    citecolor=linkpink,
    linkcolor=linkpink,
    urlcolor=linkpink
}
\usepackage{url}

\usepackage{wrapfig}
\usepackage{xcolor}

\newtheorem{proposition}{Proposition}
\newtheorem{corollary}{Corollary}

\AtBeginDocument{%
  }
\begin{document}

\title{HERO: Preserving Structure and Semantics in Heterogeneous Continual Graph Learning}


\iclrfinalcopy

\author{
Guiquan Sun$^{1}$ \qquad
Xikun Zhang$^{2,*}$ \qquad
Jingchao Ni$^{3}$ \qquad
Dongjin Song$^{1,*}$ \\
$^{1}$School of Computing, University of Connecticut \\
$^{2}$RMIT University, Melbourne \\
$^{3}$Department of Computer Science, University of Houston \\
\texttt{\{guiquan.sun,dongjin.song\}@uconn.edu} \\
\texttt{xikun.zhang@rmit.edu.au}, \texttt{jni7@uh.edu} 
}
\begingroup
\renewcommand{\thefootnote}{}
\footnotetext{$^{*}$Corresponding authors}
\addtocounter{footnote}{-1}
\endgroup

\maketitle

\begin{abstract}
Machine learning on heterogeneous graphs has advanced rapidly, driven by the diverse entities, relations, and semantics inherent in real-world data. However, most existing studies assume static graphs, whereas real-world graph data are often continuously updated. Continual learning in this setting is particularly challenging because historical knowledge is encoded not only in target-node representations, but also in their heterogeneous structural contexts and relation-dependent semantic responses. To this end, we introduce HERO, a heterogeneous continual graph learning framework that preserves historical knowledge at both structural and semantic levels while adapting to incoming tasks. We analyze experience replay as an approximation to the unavailable historical gradient and decompose its error into target-selection and context-reconstruction terms. This analysis shows that replaying target nodes alone is generally insufficient for relation-sensitive HGNNs, as removing typed neighborhood context can alter both representations and historical gradients. Motivated by this, HERO employs DiSCo to select representative target nodes and reconstruct compact heterogeneous contexts through relation-aware multi-type neighbor expansion. To further preserve knowledge that cannot be captured by replayed subgraphs alone, HERO aligns task-conditioned predictions and heterogeneous semantic responses through multi-level knowledge distillation. A lightweight look-ahead adaptation step is additionally used to improve plasticity on incoming tasks. Experiments on four datasets with three HGNN backbones show that HERO achieves the best or tied-best average performance among state-of-the-art baselines, while maintaining competitive forgetting. Our code is available at \url{https://github.com/gqBond/HCGL}.
\end{abstract}

\section{Introduction}
\label{intro}
Heterogeneous graphs model complex real-world systems through multiple node and relation types~\citep{shi2016heterogeneous}, while heterogeneous graph neural networks (HGNNs) learn expressive representations by capturing their type-specific structural and semantic interactions~\citep{sun2012mining,wang2019heterogeneous,hu2020heterogeneous}. Most existing HGNNs, however, are developed under static settings, whereas many real-world applications require models to adapt to continuously evolving heterogeneous graphs without retaining full historical observations. This gives rise to heterogeneous continual graph learning (HCGL), where an HGNN must adapt to incoming tasks while preserving knowledge acquired from previous ones.

Directly applying existing continual learning methods to HCGL overlooks how historical knowledge is distributed across heterogeneous graph computation. Existing replay methods mainly preserve selected instances, while distillation methods typically constrain model outputs. 
However, an HGNN prediction also depends on typed neighborhoods and relation-specific semantic interactions, which can change even when the target nodes or output predictions are preserved. Consequently, preserving a target node does not fully reconstruct the historical computation that produced its prediction, while preserving its output does not guarantee that the underlying relation-dependent semantics remain unchanged. We argue that effective HCGL therefore requires preserving historical knowledge at both the structural and semantic levels.

We analyze the structural challenge from an optimization perspective. When past data are unavailable, experience replay estimates the historical gradient using a limited-budget memory buffer.
We characterize the resulting increase in historical risk through the gradient approximation error, which can be decomposed into two sources: target-selection error, caused by unrepresentative replay samples of past data, and context-reconstruction error, caused by incomplete heterogeneous neighborhoods. For relation-sensitive HGNNs, we further show that target-node-only replay is generally insufficient to recover the historical prediction function, since the same target nodes can induce different representations under different typed neighborhoods. Effective heterogeneous replay must therefore preserve both representative targets and the structural contexts needed to recover their historical predictions.

Based on this observation, we propose \textbf{HERO}, a framework for preserving heterogeneous structural contexts and semantic responses in continual graph learning. HERO first introduces \textbf{DiSCo} (\underline{Di}versity \underline{S}ampling with semantic \underline{Co}nsistency), which reduces replay approximation error through representative target-node selection and relation-aware multi-type neighbor expansion, producing compact replay subgraphs that better reconstruct historical heterogeneous contexts. Yet context reconstruction does not constrain how the model uses these contexts: the roles of individual relations and semantic channels may still drift as new tasks are learned. HERO addresses this problem with \textbf{HKD} (Heterogeneity-aware Knowledge Distillation), which aligns the current model with historical teachers at two levels: task-conditioned predictions and heterogeneous semantic responses. A lightweight look-ahead step adapts the model to a representative subset of the incoming task, improving plasticity to new knowledge.

Our main contributions are summarized as follows:
\begin{enumerate}[leftmargin=.15in]
\item We provide an optimization view of heterogeneous replay, showing that replay-based retention depends on approximating the unavailable historical gradient. We decompose this approximation error into target-selection and heterogeneous-context reconstruction terms, and establish that target-node-only replay is generally insufficient for relation-sensitive HGNNs.
\item Guided by this analysis, we develop HERO to preserve historical knowledge at both structural and semantic levels. DiSCo reconstructs compact heterogeneous replay contexts, while HKD aligns task-conditioned predictions and relation-dependent semantic responses across learning stages.
\item We evaluate HERO on four heterogeneous graph benchmarks with three representative HGNN backbones and different task granularities. HERO achieves the best or tied-best average performance among strong continual graph learning baselines while maintaining competitive forgetting. Extensive analyses validate the roles of heterogeneous context reconstruction and semantic preservation.
\end{enumerate}

\section{Preliminaries}
\label{sec:pre}

We study continual node classification over a growing heterogeneous graph. The graph is revealed as a sequence of snapshots, where each snapshot is associated with the arrival of a new task. At each step, the learner receives the data of the current task and may retain only a bounded memory buffer from previous tasks. The objective is to adapt a shared HGNN to the incoming task while preserving its performance on all previously learned tasks.

\paragraph{Definition 1 (Growing Heterogeneous Graph).} A heterogeneous graph snapshot at step $t$ is denoted by $\mathcal{G}_t=(V_t,E_t,\mathcal{X}_t,\phi,\psi)$, where $V_t$ and $E_t$ are the node and edge sets, respectively, $\phi:V_t\rightarrow\mathcal{A}$ maps each node to a node type, and $\psi:E_t\rightarrow\mathcal{R}$ maps each edge to a relation type. The graph is heterogeneous if $|\mathcal{A}|+|\mathcal{R}|>2$. For each node type $\tau\in\mathcal{A}$, let $V_{\tau,t}=\{v\in V_t\mid\phi(v)=\tau\}$ and let $X_{\tau,t}\in\mathbb{R}^{|V_{\tau,t}|\times d_\tau}$ denote its type-specific feature matrix. We consider a fixed heterogeneous schema consisting of $\mathcal{A}$ and $\mathcal{R}$, while the graph grows through newly introduced node and edge instances. In particular, consecutive snapshots satisfy $V_{t-1}\subseteq V_t$ and $E_{t-1}\subseteq E_t$ for $t=2,\ldots,T$.

\paragraph{Definition 2 (Heterogeneous Continual Graph Learning).} Each snapshot $\mathcal{G}_t$ introduces a node classification task $\mathcal{T}_t$ with task-specific classes. The tasks are class-disjoint, and each task contains $K$ classes. Its target nodes are divided into training and test sets $V_t^{\mathrm{tr}}$ and $V_t^{\mathrm{te}}$, respectively. All tasks sequentially update the same HGNN encoder $h_{\varphi}$ and the $K$-way classifier $g_{\omega}$, i.e., $f_{\theta}(v,\mathcal{G}_t) = g_{\omega}(h_{\varphi}(v,\mathcal{G}_t))$ with $\theta=(\varphi,\omega)$. At evaluation, the task context $t$ is known to the evaluator and is used only to interpret the local prediction; it is not used to select task-specific model parameters. At step $t$, the model may retain a replay buffer $\mathcal{B}$ satisfying $|\mathcal{B}|\leq B$ ($B$ is the buffer size).

\section{Methodology} \label{method}
\label{Method}
\begin{figure}[t]
\centerline{
\includegraphics[width=\linewidth]{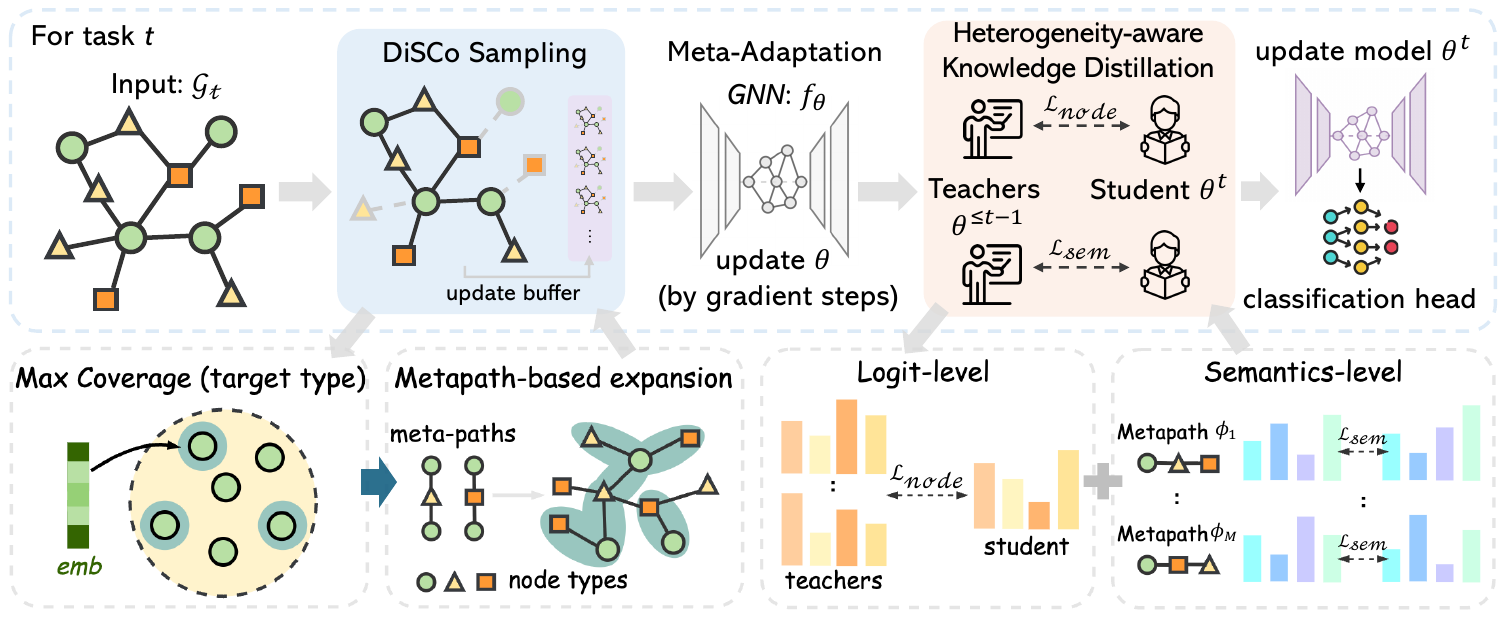}
}

\caption{
Overview of HERO for heterogeneous continual graph learning. For each incoming task, DiSCo constructs a compact replay subgraph by selecting representative target nodes and retaining their relation-aware multi-type neighborhoods, thereby preserving historical heterogeneous contexts. HKD complements replay by aligning task-conditioned predictions and heterogeneous semantic responses with historical teachers. A look-ahead meta-adaptation step further improves new-task plasticity under the joint current-task, replay, and distillation objectives.
}
\vspace{-3mm}
\label{fig:overview}
\end{figure}

In this section, we present the HERO framework (see Figure~\ref{fig:overview}). HERO addresses continual learning from two complementary aspects: retaining historical heterogeneous knowledge and maintaining plasticity to incoming tasks. We first introduce DiSCo to preserve representative heterogeneous contexts, and then present HKD to retain historical prediction and semantic responses beyond explicit replay. Finally, we describe a lightweight look-ahead adaptation mechanism that improves new-task plasticity under these historical constraints.

\subsection{Diversity Sampling with Semantic Consistency}
\label{sec:disco}
Experience replay provides a direct way to retain historical knowledge, but conventional replay strategies are not designed to preserve the heterogeneous computation contexts of HGNNs. In particular, storing representative target nodes alone may discard the typed neighborhoods and relation-specific interactions required to reproduce their historical training signals. 
To address this, we propose DiSCo (\underline{Di}versity \underline{S}ampling with semantic \underline{Co}nsistency), which jointly considers target-node diversity and heterogeneous structural context. DiSCo proceeds in two stages: (1) selecting representative target-type nodes by maximizing diversity in the feature or embedding space, and (2) expanding to other types of nodes through relation-type-aware importance estimation, ensuring that both semantic diversity and heterogeneous structural patterns are preserved.

\textbf{Replay quality as historical-gradient approximation.}
We first view experience replay as an approximation to the unavailable historical training objective. Let $F_{<t}(\theta)$ denote the full-data risk over tasks observed before $\mathcal{T}_t$. We define
\[
g_o=\nabla_\theta F_{<t}(\theta),\qquad
g_t=\nabla_\theta\mathcal{L}_{\mathcal{T}_t}(\mathcal{G}_t,\theta),\qquad
g_{\mathcal B}
=\nabla_\theta\mathcal{L}_{\mathrm{er}}(\mathcal{G}_{\mathrm{sub}},\theta).
\]
where $g_{\mathcal B}$ denote the gradient induced by the replay memory. A replay-based update can be written as
\[
\theta^+
=
\theta-\rho
\left(
g_t+\lambda_{\mathrm{er}}g_{\mathcal B}
\right),
\]
where $\rho$ is the learning rate and $\lambda_{\text{er}}$ is the trade-off coefficient.

\begin{proposition}[One-step historical-risk bound]
\label{prop:replay-bound}
Suppose $F_{<t}$ is $L_F$-smooth and define the replay-gradient approximation error as
\[
\epsilon_{\mathcal B}
=
\|g_o-g_{\mathcal B}\|_2.
\]
Then
\[
\begin{aligned}
F_{<t}(\theta^+)-F_{<t}(\theta)
\leq&
-\rho\langle g_o,g_t\rangle\\
&-\rho\lambda_{\mathrm{er}}
\left(
\|g_o\|_2^2-\|g_o\|_2\epsilon_{\mathcal B}
\right)\\
&+\frac{L_F\rho^2}{2}
\left\|
g_t+\lambda_{\mathrm{er}}g_{\mathcal B}
\right\|_2^2.
\end{aligned}
\]
\end{proposition}
The first term captures current--historical gradient interference, while the second term shows that replay imposes a stronger historical constraint when $g_{\mathcal B}$ better approximates $g_o$. Therefore, replay quality depends not only on buffer size, but also on whether the buffered graph reproduces historical training signals. To identify the sources of $\epsilon_{\mathcal B}$, let $g_{\mathcal B}^{\mathrm{orig}}$ be the hypothetical gradient obtained from the same selected target nodes while retaining their original historical graph contexts. This quantity is used only for analysis.
By the triangle inequality,
\[
\epsilon_{\mathcal B}
\leq
\underbrace{
\|g_o-g_{\mathcal B}^{\mathrm{orig}}\|_2
}_{\epsilon_{\mathrm{sel}}}
+
\underbrace{
\|g_{\mathcal B}^{\mathrm{orig}}-g_{\mathcal B}\|_2
}_{\epsilon_{\mathrm{ctx}}}.
\]
Here, $\epsilon_{\mathrm{sel}}$ measures how well the selected target
nodes represent the historical data, whereas
$\epsilon_{\mathrm{ctx}}$ measures the distortion caused by
reconstructing them with incomplete heterogeneous contexts.

\begin{proposition}[Insufficiency of target-node-only replay]
\label{prop:target-only}
For a relation-sensitive HGNN whose prediction is non-constant in at least one relation-specific message, there exist two heterogeneous graphs $\mathcal{G}$ and $\mathcal{G'}$ containing the same target-node feature and label $(x_v,y_v)$ but inducing different representations and predictions:
\[
h_\theta(v,\mathcal{G})\neq h_\theta(v,\mathcal{G'}),
\qquad
f_\theta(v,\mathcal{G})\neq f_\theta(v,\mathcal{G'}).
\]
Thus, target-node features and labels alone are not sufficient, in general, to reconstruct the historical prediction function.
\end{proposition}
Proposition~\ref{prop:target-only} establishes the need to retain heterogeneous neighborhoods. We further show in Appendix~\ref{app:disco-theory} that, under standard Lipschitz conditions,
\[
\epsilon_{\mathrm{ctx}} \leq
C \sum_{v\in\mathcal B} \sum_{r\in\mathcal R}
\|W_r\|_2 \Delta_r(v;\mathcal{G},\widetilde{\mathcal{G}}),
\]
where $C>0$ is a constant and $\Delta_r(v;\mathcal{G},\widetilde{\mathcal{G}})$ measures omitted relation-specific messages and changes in aggregation weights in replay subgraph $\widetilde{\mathcal{G}}$.
This analysis motivates the two stages of DiSCo. Coverage Maximization (CM) selects representative target nodes as a practical surrogate for reducing $\epsilon_{\mathrm{sel}}$, while multi-type neighbor expansion retains typed neighborhoods and relation-specific connections to reduce $\epsilon_{\mathrm{ctx}}$.

\textbf{Step 1: Reducing target-selection error.} The first term $\epsilon_{\mathrm{sel}}$ arises when the retained target nodes fail to represent the historical target distribution. Prior research has shown that different data points contribute unequally to model training: some significantly enhance performance, while others offer limited benefit \citep{toneva2018empirical}.
Motivated by this, we use Coverage Maximization as a practical surrogate for reducing this discrepancy by retaining class-separated target nodes in the feature (embedding) space.
Formally, given the training node set $V_t^{tr}$ of task $\mathcal{T}_t$, CM selects a subset by maximizing the coverage of the feature/embedding space:
\begin{equation}
\label{CM}
\mathcal{N}(v_i) = \{v_j \mid \text{dist}(v_i, v_j) < d, \, \mathcal{Y}(v_i) \ne \mathcal{Y}(v_j) \},
\end{equation}
where $\mathcal{Y}(v_i)$ denotes the label of node $v_i$, and $\mathcal{N}(v_i)$ represents the set of nodes from different classes that are within a distance $d$ of $v_i$. Given the memory buffer $\mathcal{B}_{\tau_t}$ for the target node type $\tau_t$, we select $e$ nodes per class with the smallest $|\mathcal{N}(v_i)|$ as representative experiences.

\textbf{Step 2: Reducing heterogeneous-context error.}
Even for a representative target set, replaying isolated target nodes can produce a large $\epsilon_{\mathrm{ctx}}$, because an HGNN prediction depends on typed neighbors and relation-specific message passing. We therefore expand each selected target node toward multiple node types and retain relation-aware neighbors. The resulting induced subgraph provides a compact approximation to the original heterogeneous computation context. Under standard \textit{Lipschitz} assumptions on message passing and the loss gradient, $\epsilon_{\mathrm{ctx}}$ can be bounded by the relation-specific message discrepancy between the original graph and the replay subgraph. The detailed result is provided in Appendix~\ref{app:context-bound}.

Let $\mathcal S_t\subseteq V_t^{\mathrm{tr}}$ denote the representative target nodes selected in Step 1, and let $\tau_t\in\mathcal A$ denote their target node type. For each non-target node type $\tau\in\mathcal A\setminus\{\tau_t\}$, let $\mathcal R(\tau_t,\tau)\subseteq\mathcal R$ denote the set of relation types connecting $\tau_t$ and $\tau$. We collect the candidate context nodes of type $\tau$ as
\begin{equation}
\label{candidate}
\mathcal C_{\tau,t}
=
\bigcup_{v\in\mathcal S_t}
\bigcup_{r\in\mathcal R(\tau_t,\tau)}
\left\{
u\in V_{\tau,t}
\;\middle|\;
(v,u)\in E_{r,t}
\lor
(u,v)\in E_{r,t}
\right\},
\end{equation}
where $V_{\tau,t}$ denotes the nodes of type $\tau$ available at task $t$, and $E_{r,t}$ denotes the edges of relation type $r$ in the current graph snapshot.
We score each candidate node by its relation-specific connectivity:
\begin{equation}
\label{importance_score}
\pi_t(v) = \sum_{r\in\mathcal R} \deg_{r,t}(v),
\end{equation}
where $\deg_{r,t}(v)$ measures the connectivity strength of relation type $r$ for node $v$ at task $t$.
For non-target node type $\tau$ with buffer budget $b_\tau$, we select top-$b_\tau$ candidate nodes by $\pi_t(v)$:
\begin{equation}
\label{topk}
\mathcal S_{\tau,t} =
\operatorname{TopK}_{v\in\mathcal C_{\tau,t}}
\left( \pi_t(v), b_\tau \right).
\end{equation}
Using the representative target set $\mathcal S_t$ and the selected context nodes
$\{\mathcal S_{\tau,t}\}_{\tau\in \mathcal A\setminus\{\tau_t\}}$, we construct the induced heterogeneous replay subgraph $\mathcal G_{\mathrm{sub}}^t=(V_t',E_t')$ as
\begin{equation}
\label{subg_cons}
V_t' = \mathcal S_t \cup
\bigcup_{\tau\in\mathcal A\setminus\{\tau_t\}} \mathcal S_{\tau,t}, \qquad
E_t' = \left\{ (u,v)\in E_t \mid u,v\in V_t' \right\}.
\end{equation}
\vspace{-1mm}
For the selected target nodes in $\mathcal S_t$, we retain their features and labels, while the sampled non-target nodes and induced edges are retained as heterogeneous message-passing context. The constructed subgraphs are stored in the replay memory. When learning task $\mathcal T_t$, the model combines the current-task loss with a replay loss evaluated on the labeled historical target nodes using their reconstructed heterogeneous contexts:
\begin{align}
\label{eq:replay}
\mathcal{L} &= \mathcal{L}_{\mathcal{T}_t}(\mathcal{G}_t, \theta) + \lambda_{er} \mathcal{L}_{er}(\mathcal{G}_{sub}, \theta)\\& = - \sum_{v_i \in V_t^{tr}} \mathcal{Y}(v_i) \log \hat{\mathcal{Y}}(v_i)
\notag
 - \lambda_{er} \sum_{v_j \in \mathcal{B}} \mathcal{Y}(v_j) \log \hat{\mathcal{Y}}(v_j),
\end{align}
where $\hat{\mathcal{Y}}(v_i)$ is the predicted label of node $v_i$ and $\lambda_{er}$ controls the contribution of buffered data in the overall loss. Equation~\ref{eq:replay} uses the DiSCo subgraph to instantiate $g_{\mathcal B}$. The two stages of DiSCo target the selection and context components of the replay approximation error, respectively.


\subsection{Heterogeneity-aware Knowledge Distillation for Task Alignment}\label{sec:hkd}

While DiSCo preserves representative heterogeneous computation contexts, replay alone cannot fully constrain the learner's semantic changes across tasks. Existing knowledge distillation methods are often developed for homogeneous scenarios and mainly focus on prediction-level alignment. For HGNNs, however, relation- and metapath-specific responses may drift even when output behavior is partially preserved.
In contrast, our HKD (\underline{H}eterogeneity-aware \underline{K}nowledge \underline{D}istillation) is explicitly tailored for HCGL. Instead of relying solely on logit-based distillation, we propose a two-level alignment strategy that leverages both prediction and semantic signals to capture graph heterogeneity. The logit-level distillation transfers soft label distributions to retain discriminative knowledge, while the semantic-level alignment matches metapath-aware attention scores, enabling the student model to preserve high-order structural patterns unique to heterogeneous graphs.

\noindent \textbf{Logit-level.}
Inspired by previous distillation-based approaches~\citep{li2017learning,tian2023knowledge}, we transfer the soft knowledge from the teacher model (previous task classifier) to the student model (current task classifier), enabling the student model to learn both new knowledge from the current task and retained knowledge from past tasks. Mimicking the teacher’s prediction results enables the student model to learn secondary information from previous tasks that cannot be expressed by the experience replay data alone. Soft knowledge from teacher model is formulated as the predicted probability of the labels in the current task data:
\begin{equation}
\label{softknowledge}
    p_{T,l}(z_i, t) = \text{Softmax}\big(f_{T,l}(z_i), t\big) = \frac{\exp \left[f_{T,l}(z_i)/t\right]}{\sum_j \exp \left[f_{T,l}(z_j)/t\right]},
\end{equation}
where $z_i$ is the embedding of node $v_i$ in $\mathcal{V}_i^{tr}$, $f_{T,i}(z_i)$ is the score logit of $z_i$ obtained from $l$-th teacher model, and $t$ is the temperature index to soften the peaky softmax distribution \citep{hinton2015distilling}. Thus, the knowledge distillation loss of the teacher model and the student model is defined as follows:
\begin{equation}
\label{logit_loss}
    \mathcal{L}_{\text{logit}} = \frac{t^2}{N} \sum_{l}^N \sum_{v_i \in \mathcal{V}_i^{tr}}  p_{T,l}(z_i, t) \log \frac{p_{T,l}(z_i, t)}{p_S(z_i, t)},
\end{equation}
where $N$ is the number of used teacher models, $p_{T,l}$ and $p_S$ are the predicted distributions of teacher model and student model respectively.

\noindent \textbf{Semantic-level.}
To preserve metapath-induced semantic patterns, we propose an attention-based structural distillation method. Given a predefined metapath set $\mathcal{P} = \{P_1, \dots, P_M\}$, let $\boldsymbol{\alpha}_{P_m}^{T}(v_i)$ and $\boldsymbol{\alpha}_{P_m}^{S}(v_i)$ denote the attention coefficients for metapath $P_m$ computed by the teacher and student models, respectively. These attention coefficients reflect the importance of different semantic contexts encoded by heterogeneous structural patterns. To align the semantic-level knowledge between the teacher and the student, we define the semantic alignment loss as:
\begin{equation}
\label{sem_loss}
\mathcal{L}_{\text{sem}} = \frac{1}{N}\sum_{l=1}^N \sum_{m=1}^M \| \boldsymbol{\alpha}_{l,P_m}^{T} - \boldsymbol{\alpha}_{P_m}^{S} \|_2
\end{equation}
where $\boldsymbol{\alpha}_{P_m} \in \mathbb{R}^{|V^{tr}|}$ is the normalized attention vector over all nodes for metapath ${P_m}$. 
We provide a detailed discussion in Appendix~\ref{app:extension} on how this module can be extended to arbitrary HGNNs beyond metapath-based architectures.

This loss guides the student model to capture relational importance and structural dependencies aligned with the teacher model, thereby preserving semantic information that is critical in heterogeneous graphs. Afterwards, we combine the logit level loss and the semantic level loss:
\begin{equation} 
\label{kd_loss}
    \mathcal{L}_{kd} = \lambda_{\text{logit}}\mathcal{L}_{\text{logit}} + \lambda_{\text{sem}}\mathcal{L}_{\text{sem}},
\end{equation}
where $\lambda_{\text{logit}}$ and $\lambda_{\text{sem}}$ are both trade-off weights for balancing the losses.
The final objective of node representation learning is to minimize the joint loss including current task loss $\mathcal{L}_{\mathcal{T}_i}$, experience replay loss $\mathcal{L}_{er}$, and the KD loss $\mathcal{L}_{kd}$:
\begin{equation} 
\label{eq:joint_loss}
    \mathcal{L}_{\text{meta}} = \mathcal{L}_{\mathcal{T}_t} + \lambda_{er} \mathcal{L}_{er} + \lambda_{kd} \mathcal{L}_{kd},
\end{equation}
where $\lambda_{kd}$ is a trade-off weight for balancing the losses. 

\subsection{Fast Adaptation with Gradient-based Meta-Learning}\label{sec:lamaml}
The previous two components primarily focus on retaining historical
knowledge. We now consider the complementary problem of maintaining
plasticity when a new task arrives.
Inspired by the look-ahead optimization principle of La-MAML~\citep{gupta2020look}, we adopt a simplified meta-adaptation strategy tailored to heterogeneous graphs. The main idea is to first perform a fast update on a small representative support graph and then use the joint continual objective to regulate the adaptation direction.
Specifically, for the current task $\mathcal{T}_t$, we construct a support graph $\mathcal{G}_{\mathrm{sup}}^t$ from $e$ representative nodes selected by the sampling method introduced in Section~\ref{sec:disco}. 
At optimization step $j$, we obtain the fast-adapted parameters through
\begin{equation}
\theta_{\mathrm{fast}}^j
=
\theta^j
-
\alpha^j \odot
\nabla_{\theta^j}
\mathcal{L}_{\mathrm{sup}}
(\mathcal{G}_{\mathrm{sup}}^t,\theta^j),
\end{equation}
where $\alpha^j$ denotes the parameter-wise adaptation rate.

Rather than directly accepting this fast update, we evaluate it using the joint continual learning objective $\mathcal{L}_{\mathrm{meta}}$ in Equation~\ref{eq:joint_loss}. Following the look-ahead principle, the adaptation rates are updated as
\begin{equation}
\alpha^{j+1}
=
\alpha^j
-
\eta \odot
\nabla_{\alpha^j}
\mathcal{L}_{\mathrm{meta}}
(\mathcal{G}_t,\theta_{\mathrm{fast}}^j),
\end{equation}
and then update the model parameters as
\begin{equation}
\theta^{j+1}
=
\theta_{\mathrm{fast}}^j
-
\max(0,\alpha^{j+1})
\odot
\nabla_{\theta_{\mathrm{fast}}^j}
\mathcal{L}_{\mathrm{meta}}
(\mathcal{G}_t,\theta_{\mathrm{fast}}^j).
\end{equation}

This look-ahead procedure complements the two retention mechanisms introduced above. The support graph provides an informative direction for adapting to the incoming task, while the joint objective evaluates whether this fast update remains compatible with the historical constraints supplied by DiSCo and HKD. Compared with directly optimizing the joint loss, the parameter-wise adaptation rates allow the update magnitude to be adjusted according to this compatibility, providing a lightweight mechanism for maintaining new-task plasticity while respecting historical knowledge.
The complete training procedure of HERO is summarized in Algorithm~\ref{alg:HERO}.

 \section{Experiments} \label{exp}

\begin{table}[t]
\centering
\small
\caption{Performance comparison on four benchmark datasets using HAN as the backbone. Results on Yelp are reported under the 5 tasks/3-way setting.
The symbol $\uparrow$~($\downarrow$) indicates that higher~(lower) is better.
The best results are highlighted in \textbf{bold}, while the second-best results are \underline{underlined}.}
\setlength\tabcolsep{1mm}
\resizebox{0.95\linewidth}{!}{%
\begin{tabular}{l|cc|cc|cc|cc}
\toprule
\multirow{2}{*}{Methods}
& \multicolumn{2}{c|}{DBLP}
& \multicolumn{2}{c|}{IMDB}
& \multicolumn{2}{c|}{Freebase}
& \multicolumn{2}{c}{Yelp} \\
\cmidrule(lr){2-3}
\cmidrule(lr){4-5}
\cmidrule(lr){6-7}
\cmidrule(lr){8-9}
& AP / \% $\uparrow$ & AF / \% $\downarrow$
& AP / \% $\uparrow$ & AF / \% $\downarrow$
& AP / \% $\uparrow$ & AF / \% $\downarrow$
& AP / \% $\uparrow$ & AF / \% $\downarrow$ \\
\midrule

Finetune
& 82.8 $\pm$ 4.6 & 26.6 $\pm$ 9.1
& 68.8 $\pm$ 2.2 & 26.0 $\pm$ 2.8
& 53.8 $\pm$ 4.7 & 25.7 $\pm$ 8.8
& 24.5 $\pm$ 0.5 & 44.8 $\pm$ 1.6 \\

Joint Train
& 95.2 $\pm$ 0.6 & 1.7 $\pm$ 1.0
& 80.4 $\pm$ 0.7 & 3.9 $\pm$ 0.8
& 65.6 $\pm$ 0.9 & 2.8 $\pm$ 2.0
& 63.0 $\pm$ 2.3 & 0.0 $\pm$ 1.6 \\

\midrule

EWC
& 85.4 $\pm$ 5.5 & 21.5 $\pm$ 10.6
& 69.6 $\pm$ 1.2 & 25.8 $\pm$ 2.4
& 58.3 $\pm$ 2.4 & 18.4 $\pm$ 4.2
& 25.3 $\pm$ 2.1 & 44.8 $\pm$ 0.9 \\

MAS 
& 91.1 $\pm$ 0.8 & 8.3 $\pm$ 1.5
& 72.0 $\pm$ 2.3 & 20.5 $\pm$ 3.9
& 59.0 $\pm$ 1.0 & 14.8 $\pm$ 0.9
& 36.7 $\pm$ 10.9 & 27.6 $\pm$ 16.2 \\

TWP 
& 90.5 $\pm$ 1.3 & 10.1 $\pm$ 2.6
& 74.8 $\pm$ 2.7 & 14.5 $\pm$ 4.4
& \underline{60.9 $\pm$ 4.3} & \textbf{10.8 $\pm$ 6.9}
& 32.0 $\pm$ 12.1 & 36.4 $\pm$ 15.1 \\

ER-GNN 
& 89.7 $\pm$ 2.9 & 13.1 $\pm$ 6.1
& \underline{74.9 $\pm$ 2.3} & \underline{12.7 $\pm$ 4.2}
& 56.5 $\pm$ 1.3 & 19.6 $\pm$ 0.9
& 40.2 $\pm$ 14.2 & 22.5 $\pm$ 18.2 \\

MetaCLGraph
& 89.9 $\pm$ 0.9 & 11.4 $\pm$ 2.3
& 74.8 $\pm$ 3.2 & 14.8 $\pm$ 5.7
& 59.4 $\pm$ 2.5 & 13.0 $\pm$ 4.1
& 37.4 $\pm$ 12.3 & 30.7 $\pm$ 15.9 \\

FTF-ER 
& 90.8 $\pm$ 1.4 & 9.8 $\pm$ 2.4
& 72.0 $\pm$ 4.2 & 19.7 $\pm$ 6.4
& 58.1 $\pm$ 0.7 & 14.6 $\pm$ 0.6
& 32.8 $\pm$ 12.5 & 36.2 $\pm$ 16.0 \\

DMSG 
& \underline{93.5 $\pm$ 0.6} & \textbf{0.7 $\pm$ 0.6}
& 68.8 $\pm$ 3.9 & 25.1 $\pm$ 7.8
& 57.1 $\pm$ 2.4 & 17.5 $\pm$ 3.6
& \underline{50.3 $\pm$ 0.6} & \textbf{5.6 $\pm$ 0.5} \\

\rowcolor{gray!20}
HERO
& \textbf{93.6 $\pm$ 0.8} & \underline{4.3 $\pm$ 1.6}
& \textbf{78.4 $\pm$ 1.7} & \textbf{7.3 $\pm$ 3.0}
& \textbf{60.9 $\pm$ 1.5} & \underline{11.1 $\pm$ 1.6}
& \textbf{55.9 $\pm$ 0.5} & \underline{7.5 $\pm$ 1.2} \\

\bottomrule
\end{tabular}}
\vspace{-2mm}
\label{tab:main_result}
\end{table}

\begin{figure}[t]
    \centering
    \includegraphics[width=\linewidth]{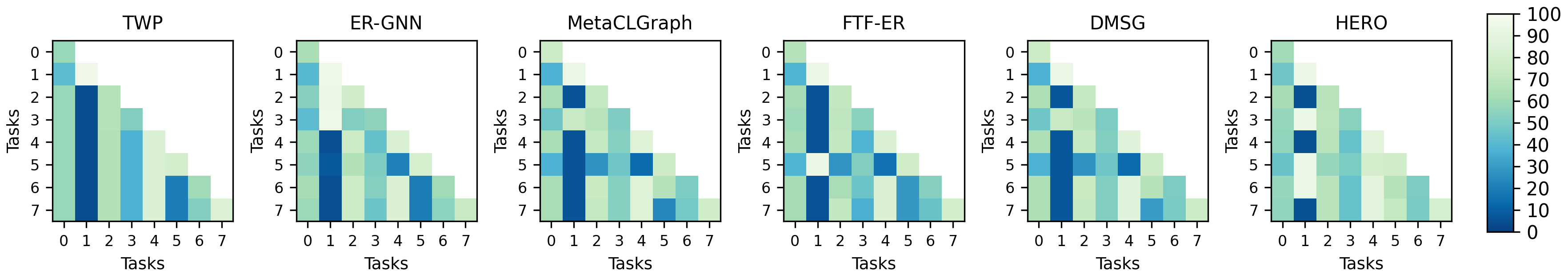}
    \vspace{-5mm}
    \caption{Accuracy matrix of selected continual graph learning methods on the Yelp dataset.}
    \vspace{-3mm}
    \label{fig:heatmap}
\end{figure}

\noindent \textbf{Experiment Setup.} In this section, we describe the experiments we perform to validate our proposed method. We evaluate HERO on four heterogeneous graph benchmarks: Yelp~\citep{yang2020heterogeneous}, IMDB, DBLP, and Freebase~\citep{lv2021we}. Following the continual-learning protocol, we partition their target classes into sequential tasks and construct the corresponding growing graph snapshots. This results in 5, 2, 2, and 3 tasks for Yelp, IMDB, DBLP, and Freebase, respectively. For Yelp, which contains a larger number of target classes, we consider two settings to examine different sequence granularities: 5 tasks/3-way and 8 tasks/2-way.
We adopt three widely used HGNN backbones in our experiments: HAN~\citep{wang2019heterogeneous}, MAGNN~\citep{fu2020magnn}, and HGT~\citep{hu2020heterogeneous}.
For baselines, we select multiple methods from the Continual Graph Learning Benchmark (CGLB) \citep{zhang2022cglb}, including EWC \citep{kirkpatrick2017overcoming}, MAS \citep{Aljundi_2018_ECCV}, ER-GNN \citep{zhou2021overcoming}, and TWP \citep{liu2021overcoming}. 
Additionally, we include several memory-replay methods: MetaCLGraph \citep{unal2023meta}, FTF-ER \citep{pang2024ftf} and DMSG~\citep{qiao2025towards}. 
We use Finetune (without any continual learning techniques) as a lower bound, and Joint Train (trained jointly on data from all tasks) as an approximate upper bound \citep{caruana1997multitask}.
For evaluation, we first construct an accuracy matrix $W^{Acc} \in \mathcal{R}^{T\times T}$, where $W^{Acc}_{i,j} (i \geq j)$ denotes the accuracy on the $j$-th task after learning task $i$. To summarize the final performance after learning all tasks, we report the Average Performance (AP): $\mathrm{AP} = \frac{1}{T} \sum_{i=1}^{T} W_{T,i}^{\mathrm{Acc}}$ and the Average Forgetting (AF): $\mathrm{AF} = \frac{1}{T-1} \sum_{i=1}^{T-1} (W_{i,i}^{\mathrm{Acc}} - W_{T,i}^{\mathrm{Acc}})$. For a more detailed description of the experimental setup, please refer to Appendix~\ref{app:config}.

\begin{wrapfigure}{l}{0.5\columnwidth}
    \centering
    \vspace{-2mm}
    \includegraphics[width=\linewidth]{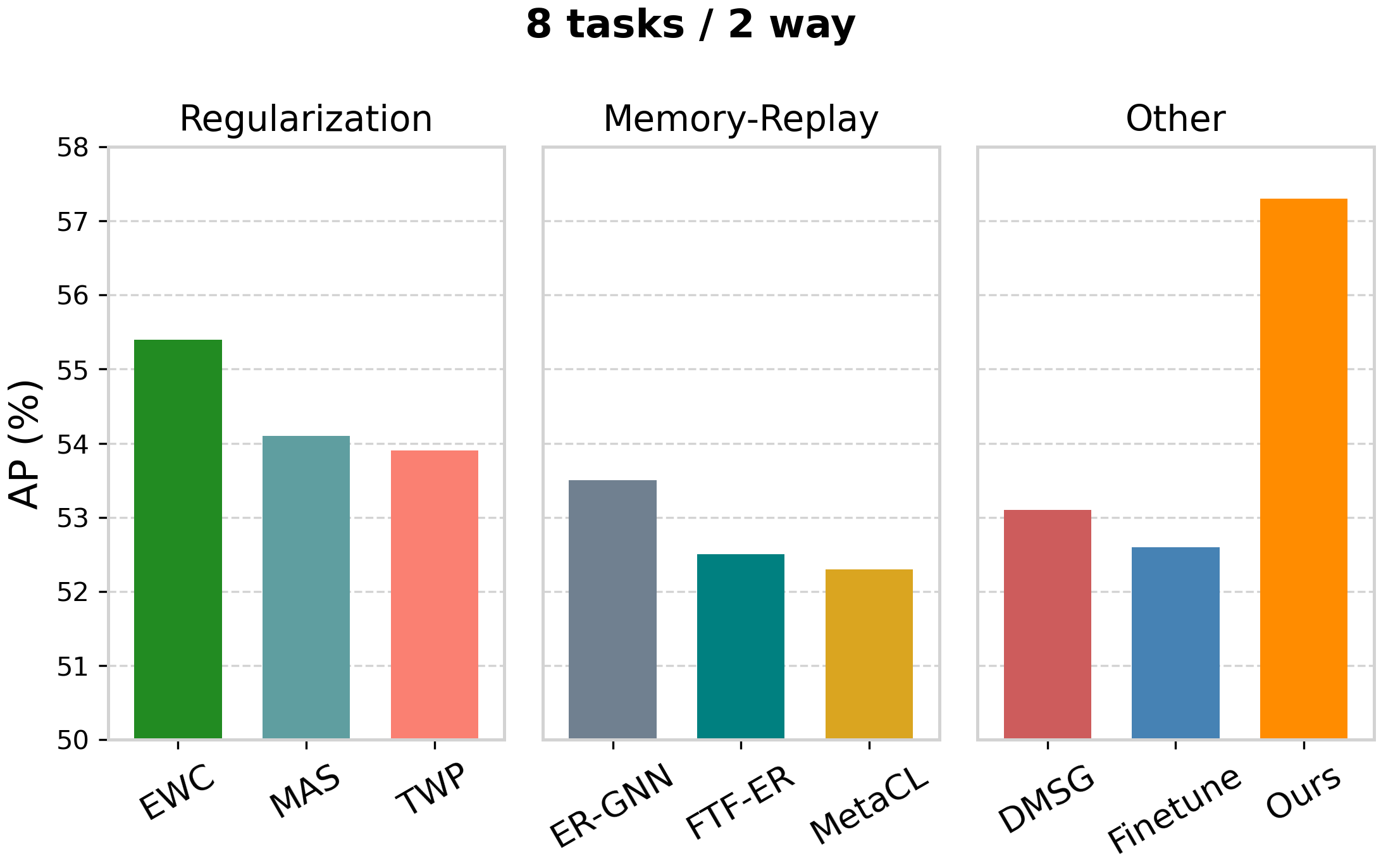}\vspace{-2mm}
    \caption{Performance comparison on Yelp under the 8-task/2-way setting.}
    \label{fig:yelp_results}
    \vspace{-2mm}
\end{wrapfigure}

\noindent \textbf{Performance Evaluation.}
The overall performance on four heterogeneous graph datasets using HAN as the backbone is reported in Table~\ref{tab:main_result}. Results with additional HGNN backbones are provided in Appendix~\ref{app:backbone}. Task-wise accuracy retention results on Yelp are further illustrated in Figure~\ref{fig:yelp_results} and Figure~\ref{fig:heatmap}. 
In addition, we conduct extended experiments to assess the robustness and scalability of HERO, including GPU memory usage and the teacher-storage cost over longer task sequences, with detailed results provided in Appendix~\ref{app:scale} and \ref{app:exp_robust}.
We highlight the following key observations:
\textbf{(1)} Sequentially fine-tuning HGNN backbones without CL mechanisms leads to substantial forgetting. For example, the Finetune baseline yields an AF of 26.6\% on DBLP with HAN, while similar forgetting is observed with the other backbones (Appendix~\ref{app:backbone}). This confirms that heterogeneous graph models are not immune to catastrophic forgetting under sequential task updates.
\textbf{(2)} Across the reported backbones, HERO achieves the highest or tied-highest AP among the continual learning baselines. The gains are particularly clear on IMDB and Yelp, while the results on DBLP and Freebase are also competitive. With HAN, HERO achieves the lowest or second-lowest AF among continual-learning baselines on all four datasets.
Additional experiments with randomized task orders on HAN and analyses using the Forgetting-aware Gap metric indicate that HERO generally maintains its advantages under task-sequence variations and often attains a favorable stability–plasticity trade-off.
\textbf{(3)} Some replay-based methods obtain lower AF than HERO in individual settings despite having lower AP. For example, DMSG achieves lower AF on DBLP and Yelp, whereas HERO obtains higher AP. This indicates that different methods operate at different points of the stability--plasticity trade-off: stronger retention does not necessarily lead to higher overall performance. HERO aims for a balanced operating point, achieving consistently strong AP while keeping forgetting competitive.


\begin{figure}[t]
\centering
\vspace{-2mm}
\includegraphics[width=0.98\linewidth]{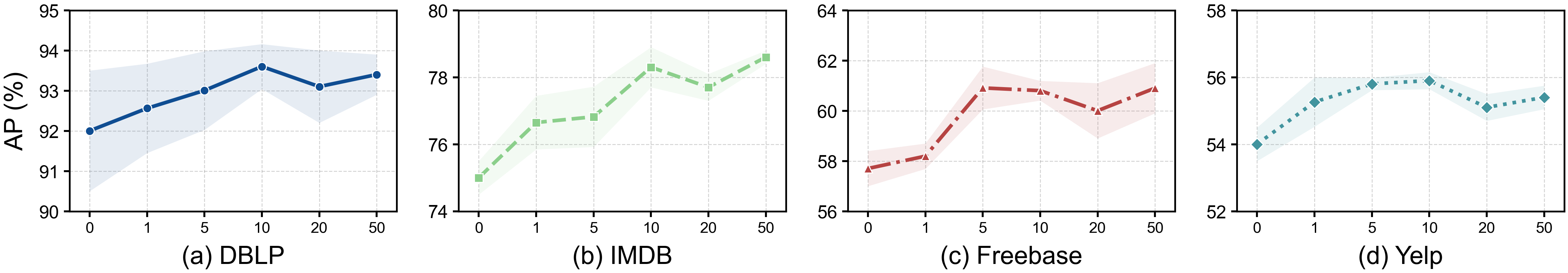}\vspace{-2mm}
\caption{Effect of the shot number used by G-MM across four datasets.} 
\vspace{-2mm}
\label{fig:shot}
\end{figure}

\begin{table}
\caption{Ablation results of HERO. We remove Experience Replay (ER), Heterogeneity-aware Knowledge Distillation (HKD), and the Gradient-based Meta-learning Module (G-MM) individually. And ``w/o DiSCo'' denotes replacing the proposed \textit{Diversity Sampling with semantic Consistency} with Coverage Maximization applied to all node types.}
\centering
\setlength\tabcolsep{1.1mm}
\resizebox{0.95\linewidth}{!}{%
\begin{tabular}{l|cc|cc|cc|cc}
\toprule
\multirow{2}{*}{Methods} & \multicolumn{2}{c|}{DBLP} & \multicolumn{2}{c|}{IMDB} & \multicolumn{2}{c|}{Freebase} & \multicolumn{2}{c}{Yelp} \\ \cline{2-9}
  & AP / \% $\uparrow$ & AF / \% $\downarrow$ & AP / \% $\uparrow$ & AF / \% $\downarrow$ & AP / \% $\uparrow$ & AF / \% $\downarrow$ & AP / \% $\uparrow$ & AF / \% $\downarrow$ \\ 
\toprule
\rowcolor{gray!20}
HERO & \textbf{93.6 $\pm$ 0.8} & \textbf{4.3 $\pm$ 1.6} & \textbf{78.4 $\pm$ 1.7} & \textbf{7.3 $\pm$ 3.0} & \textbf{60.9 $\pm$ 1.5} & \textbf{11.1 $\pm$ 1.6} & \textbf{55.9 $\pm$ 0.6} & \textbf{7.5 $\pm$ 1.2} \\
w/o ER & 89.7 $\pm$ 0.6 & 12.0 $\pm$ 0.7 & 75.1 $\pm$ 0.9 & 12.8 $\pm$ 1.7 & 58.6 $\pm$ 1.2 & 16.3 $\pm$ 2.2 & 53.9 $\pm$ 2.4 & 6.8 $\pm$ 2.4 \\
w/o HKD & 89.6 $\pm$ 1.4 & 12.1 $\pm$ 2.6 & 72.2 $\pm$ 1.5 & 19.3 $\pm$ 2.3 & 59.4 $\pm$ 1.1 & 14.7 $\pm$ 1.8 & 44.9 $\pm$ 14.3 & 19.3 $\pm$ 17.0 \\
w/o G-MM & 92.6 $\pm$ 1.7 & 5.2 $\pm$ 0.9 & 75.0 $\pm$ 2.7 & 13.3 $\pm$ 4.5 & 59.7 $\pm$ 0.9 & 12.0 $\pm$ 2.2 & 55.6 $\pm$ 1.9 & 8.0 $\pm$ 2.0 \\
w/o DiSCo & 92.4 $\pm$ 1.4 & 6.0 $\pm$ 3.1 & 75.4 $\pm$ 2.2 & 12.1 $\pm$ 5.0 & 56.2 $\pm$ 0.4 & 18.0 $\pm$ 0.5 & 45.7 $\pm$ 14.0 & 23.7 $\pm$ 15.6 \\
\bottomrule
\end{tabular}
}
\vspace{-2mm}
\label{tab:ablation}
\end{table}

\noindent \textbf{Ablation Study.}
To evaluate the effectiveness of individual modules in HERO, we conducted comprehensive ablation experiments using HAN as the backbone network across four datasets. The results are presented in Table~\ref{tab:ablation}.
We first examined the impact of each component by sequentially removing Experience Replay (ER), Heterogeneity-aware Knowledge Distillation (HKD), and the Gradient-based Meta-learning Module (G-MM) while keeping other components intact.
Removing every component decreases AP on all four datasets, while most ablations also increase AF. An exception occurs on Yelp, where removing ER slightly decreases AF from 7.5\% to 6.8\% but also reduces AP from 55.9\% to 53.9\%. This illustrates that AP and AF capture different aspects of the stability--plasticity trade-off.
The ablations reveal different roles of the three components. Removing G-MM causes a relatively smaller AP reduction (0.3 to 3.4 percentage points), consistent with its role as a lightweight adaptation component. In contrast, removing either HKD or DiSCo leads to larger degradation on several datasets. HKD has a particularly strong effect on DBLP and IMDB, whereas DiSCo causes larger forgetting increases on Freebase and Yelp. These results support the complementary roles of semantic-response preservation and heterogeneous-context replay in historical knowledge retention.
Replacing DiSCo with Coverage Maximization consistently lowers AP and generally increases AF, with particularly large degradation on Freebase and Yelp. This result is consistent with our analysis that retaining typed neighborhood context improves replay quality beyond target-node diversity alone.
Detailed analyses of HKD submodules are provided in Appendix~\ref{app:hkd_abl}.

\noindent \textbf{Hyperparameters Analysis.}
We analyze the replay buffer size, shot number (i.e., the number of support samples used by G-MM), and the distillation loss weight $\lambda_{kd}$. 
Due to space limitations, we visualize only the effect of the shot number in Figure~\ref{fig:shot} and provide the remaining results in Appendix~\ref{app:vis}.
The experimental findings are as follows:
\noindent\textbf{Distillation Loss Weight ($\lambda_{kd}$).} The preferred distillation strength is dataset dependent. IMDB benefits from relatively larger $\lambda_{kd}$, whereas Freebase performs better with smaller values around $0.1$ to $0.3$. Yelp also shows stronger performance under relatively larger distillation weights. Overall, these results indicate that the balance between current-task learning and historical constraints varies with the dataset.
\noindent \textbf{Buffer Size.}: AP varies non-monotonically with the replay budget. Moderate buffers perform best on DBLP, IMDB, and Freebase, whereas Yelp remains relatively stable for buffer sizes of 10, 20, and 100. Thus, increasing the buffer does not guarantee higher AP, and the preferred budget depends on the dataset.
\noindent \textbf{Shot Number.}: The number of sampled nodes impacts the stability-plasticity trade-off. As shown in Figure~\ref{fig:shot}, model performance steadily improves up to 10-shot, but declines at 20-shot, likely due to overfitting to the current task’s data patterns. The optimal shot number is 10, balancing historical information retention and computational efficiency.

\vspace{-2mm}
\section{Related Work}
\label{sec:re_works}
\vspace{-2mm}
\textbf{Heterogeneous Graph Neural Networks.} HGNNs model type-dependent structures and semantics through either metapath-based aggregation~\citep{wang2019heterogeneous,fu2020magnn,zhang2019heterogeneous} or metapath-free relation modeling~\citep{hu2020heterogeneous,lv2021we,zhu2019relation}. These models are commonly studied in offline settings where the training graph is jointly accessible. In contrast, our work studies how existing HGNN backbones can be sequentially updated under limited access to historical data. 
\textbf{Continual Graph Learning.} Existing CGL methods mainly rely on parameter regularization~\citep{kirkpatrick2017overcoming,Aljundi_2018_ECCV,liu2021overcoming,qiu2025efficient}, experience replay~\citep{zhou2021overcoming,choi2024dslr,zhang2024topology,qiao2025towards,song2025exploring}, or knowledge distillation~\citep{Xu2020GraphSAIL,dong2021rkd,tian2026dingle}. Many of these methods are designed for homogeneous graphs or do not explicitly preserve heterogeneous computation contexts~\citep{zhang2023ricci,guo2026graphkeeper}. In HGNNs, a target-node prediction depends on typed neighborhoods and relation- or metapath-specific semantic channels. Consequently, replaying isolated nodes may not reconstruct historical training signals, while output-level alignment alone may not preserve internal heterogeneous semantics. HERO addresses these limitations through heterogeneous subgraph replay and multi-level knowledge distillation. 
\textbf{Meta-Learning.} Optimization-based meta-learning supports rapid task adaptation by learning transferable initializations or update rules~\citep{finn2017model,nichol2018reptile,gupta2020look}. It has also been applied to continual learning and graph learning~\citep{javed2019meta,unal2023meta,zhang2022hg}. HERO adopts a lightweight look-ahead adaptation mechanism to improve plasticity on incoming tasks, while replay and distillation provide
historical constraints.

\vspace{-2mm}
\section{Conclusion}\label{sec:con}
\vspace{-2mm}
In this work, we studied heterogeneous continual graph learning, where HGNNs learn sequential tasks without full access to historical data. We showed that replaying target nodes alone can be insufficient because historical predictions depend on typed neighborhood contexts and relation-dependent semantics. Based on this observation, we proposed HERO to preserve historical knowledge at both structural and semantic levels. DiSCo selects representative target nodes and reconstructs compact heterogeneous contexts through relation-aware multi-type expansion, while HKD aligns task-conditioned predictions and semantic responses across learning stages. A lightweight look-ahead adaptation step further supports plasticity on incoming tasks. Experiments on four datasets with three representative HGNN backbones show that HERO achieves the best or tied-best average performance among strong baselines while maintaining competitive forgetting, supporting the importance of jointly preserving structure and semantics in HCGL.

\section{AI Usage Disclosure}\label{sec:AI_use}
The generative AI tools employed in this work are solely used for language polishing and refinement. These tools do not contribute to the development of the core ideas, methodologies, or results presented in the work.

\bibliographystyle{iclr2027_conference}
\bibliography{reference}

\clearpage
\appendix
\section{Methodology}

\subsection{Theoretical Analysis of DiSCo} \label{app:disco-theory} \paragraph{Proof of Proposition~\ref{prop:replay-bound}}

Since $F_{<t}$ is $\beta$-smooth, for any parameter displacement
$d$,
\[
F_{<t}(\theta+d)
\leq
F_{<t}(\theta)
+
\left\langle
\nabla_{\theta}F_{<t}(\theta),
d
\right\rangle
+
\frac{\beta}{2}\|d\|_2^2.
\]
For the replay-based update,
\[
d
=
-\eta
\left(
g_t+\lambda_{\mathrm{er}}g_{\mathcal{B}}
\right).
\]
Substituting $d$ and using
$g_o=\nabla_{\theta}F_{<t}(\theta)$ gives
\[
\begin{aligned}
F_{<t}(\theta^{+})-F_{<t}(\theta)
\leq&
-\eta
\left\langle
g_o,g_t+\lambda_{\mathrm{er}}g_{\mathcal{B}}
\right\rangle\\
&+
\frac{\beta\eta^2}{2}
\left\|
g_t+\lambda_{\mathrm{er}}g_{\mathcal{B}}
\right\|_2^2\\
=&
-\eta\langle g_o,g_t\rangle
-\eta\lambda_{\mathrm{er}}
\langle g_o,g_{\mathcal{B}}\rangle\\
&+
\frac{\beta\eta^2}{2}
\left\|
g_t+\lambda_{\mathrm{er}}g_{\mathcal{B}}
\right\|_2^2.
\end{aligned}
\]

By the definition
\[
\epsilon_{\mathcal{B}}
=
\|g_{\mathcal{B}}-g_o\|_2,
\]
we have
\[
\begin{aligned}
\langle g_o,g_{\mathcal{B}}\rangle
&=
\langle g_o,g_o\rangle
+
\langle g_o,g_{\mathcal{B}}-g_o\rangle\\
&\geq
\|g_o\|_2^2
-
\|g_o\|_2
\|g_{\mathcal{B}}-g_o\|_2\\
&=
\|g_o\|_2^2
-
\|g_o\|_2\epsilon_{\mathcal{B}},
\end{aligned}
\]
where the inequality follows from Cauchy--Schwarz. Substituting this
lower bound into the previous expression proves the result.

\paragraph{Proof of Proposition~\ref{prop:target-only}: Target-Node Insufficiency}

Consider a target node $v$ with feature $x_v$ and label $y_v$.
Let
\[
b=W_0x_v
\]
denote its self-feature contribution.

Because the HGNN is relation-sensitive, there exist a relation $r$, a node feature $x_u$, and a nonzero aggregation coefficient $a_{vu}^{r}$ such that the relation-specific message
\[
m
=
a_{vu}^{r}W_rx_u
\]
changes the target-node prediction:
\[
f_{\theta}
\left(
\sigma(b+m)
\right)
\neq
f_{\theta}
\left(
\sigma(b)
\right).
\]

Construct graph $\mathcal{G}$ by connecting $v$ to node $u$ through relation $r$. Construct graph $\mathcal{G}'$ using the same target node $v$, with the same feature $x_v$ and label $y_v$, but without this relation-specific neighbor. Then
\[
h_v(\mathcal{G})
=
\sigma(b+m),
\qquad
h_v(\mathcal{G}')
=
\sigma(b),
\]
and consequently
\[
f_{\theta}(v,\mathcal{G})
\neq
f_{\theta}(v,\mathcal{G}').
\]

A memory that stores only $(x_v,y_v)$ is identical for $\mathcal{G}$ and $\mathcal{G}'$, and therefore cannot determine which historical prediction function should be reconstructed.

\paragraph{Relation-Context Distortion under Subgraph Replay}\label{app:context-bound}

Let $\widetilde{\mathcal{G}}\subseteq \mathcal{G}$ be a replay subgraph. For relation
$r$, define
\[
\mathcal{C}_r(v)
=
\mathcal{N}_r^\mathcal{G}(v)
\cap
\mathcal{N}_r^{\widetilde{\mathcal{G}}}(v)
\]
as the retained neighbors, and
\[
\mathcal{O}_r(v)
=
\mathcal{N}_r^\mathcal{G}(v)
\setminus
\mathcal{N}_r^{\widetilde{\mathcal{G}}}(v)
\]
as the omitted neighbors. We define the relation-context discrepancy as
\[
\begin{aligned}
\Delta_r(v;\mathcal{G},\widetilde{\mathcal{G}})
=&
\sum_{u\in\mathcal{C}_r(v)}
\left|
a_{vu}^{r,\mathcal{G}}
-
a_{vu}^{r,\widetilde{\mathcal{G}}}
\right|\\
&+
\sum_{u\in\mathcal{O}_r(v)}
\left|
a_{vu}^{r,\mathcal{G}}
\right|.
\end{aligned}
\]
The first term accounts for changes in normalized aggregation
weights on retained neighbors, while the second term accounts for
omitted relation-specific messages.

Let the pre-activation representation be
\[
s_v(\mathcal{G})
=
W_0x_v
+
\sum_{r\in\mathcal{R}}
\sum_{u\in\mathcal{N}_r^G\mathcal{G}(v)}
a_{vu}^{r,\mathcal{G}}W_rx_u.
\]
Because $\widetilde{\mathcal{G}}$ preserves the same target node,
the self-feature terms cancel. Hence,
\[
\begin{aligned}
s_v(\mathcal{G})-s_v(\widetilde{\mathcal{G}})
=&
\sum_{r\in\mathcal{R}}
\sum_{u\in\mathcal{C}_r(v)}
\left(
a_{vu}^{r,\mathcal{G}}
-
a_{vu}^{r,\widetilde{\mathcal{G}}}
\right)W_rx_u\\
&+
\sum_{r\in\mathcal{R}}
\sum_{u\in\mathcal{O}_r(v)}
a_{vu}^{r,\mathcal{G}}W_rx_u.
\end{aligned}
\]

Using the triangle inequality, the operator norm inequality, and
$\|x_u\|_2\leq H$, we obtain
\[
\begin{aligned}
\|s_v(\mathcal{G})-s_v(\widetilde{\mathcal{G}})\|_2
\leq&
H
\sum_{r\in\mathcal{R}}
\|W_r\|_2
\Delta_r(v;\mathcal{G},\widetilde{\mathcal{G}}).
\end{aligned}
\]

Since $\sigma$ is $L_{\sigma}$-Lipschitz,
\[
\begin{aligned}
\|h_v(\mathcal{G})-h_v(\widetilde{\mathcal{G}})\|_2
&=
\|\sigma(s_v(\mathcal{G}))-\sigma(s_v(\widetilde{\mathcal{G}}))\|_2\\
&\leq
L_{\sigma}
\|s_v(\mathcal{G})-s_v(\widetilde{\mathcal{G}})\|_2\\
&\leq
L_{\sigma}H
\sum_{r\in\mathcal{R}}
\|W_r\|_2
\Delta_r(v;\mathcal{G},\widetilde{\mathcal{G}}).
\end{aligned}
\]

\begin{corollary}[Context-induced replay-gradient error]
Suppose the per-example replay-gradient map
$\psi(h,y)$ is $L_g$-Lipschitz in $h$. For a fixed set of replay
targets $\mathcal{S}$,
\[
\begin{aligned}
&
\left\|
\frac{1}{|\mathcal{S}|}
\sum_{v\in\mathcal{S}}
\psi(h_v(\mathcal{G}),y_v)
-
\frac{1}{|\mathcal{S}|}
\sum_{v\in\mathcal{S}}
\psi(h_v(\widetilde{\mathcal{G}}),y_v)
\right\|_2\\
&\quad\leq
\frac{L_gL_{\sigma}H}{|\mathcal{S}|}
\sum_{v\in\mathcal{S}}
\sum_{r\in\mathcal{R}}
\|W_r\|_2
\Delta_r(v;\mathcal{G},\widetilde{\mathcal{G}}).
\end{aligned}
\]
\end{corollary}

By the triangle inequality and the Lipschitz continuity of $\psi$,
\[
\begin{aligned}
&
\left\|
\frac{1}{|\mathcal{S}|}
\sum_{v\in\mathcal{S}}
\left[
\psi(h_v(\mathcal{G}),y_v)
-
\psi(h_v(\widetilde{\mathcal{G}}),y_v)
\right]
\right\|_2\\
&\leq
\frac{L_g}{|\mathcal{S}|}
\sum_{v\in\mathcal{S}}
\|h_v(\mathcal{G})-h_v(\widetilde{\mathcal{G}})\|_2.
\end{aligned}
\]
Substituting the relation-context distortion bound proves the result.

\subsection{Detailed Algorithm}\label{app:alg}

To better understand the proposed method, we provide the detailed algorithm~\ref{alg:HERO} to illustrate the learning process.
\begin{algorithm}[h]
\caption{The procedure of HERO}
\label{alg:HERO}
\begin{algorithmic}[1]
\State \textbf{Input:} Continual tasks $\mathcal{T}=\{\mathcal{T}_1,\ldots,\mathcal{T}_T\}$; buffer $\mathcal{B}$; number of representative nodes $e$; adaptation rates $\alpha$; meta learning rate $\eta$; sampling algorithm $\operatorname{DiSCo}(\cdot,\cdot)$.
\State \textbf{Output:} Continually learned model $f_{\theta}$.
\State Initialize model parameters $\theta$, adaptation rates $\alpha$, buffer $\mathcal{B}\leftarrow\emptyset$, and teacher set $\mathcal{F}\leftarrow\emptyset$

\For{$t=1$ {\bfseries to} $T$}
    \State Obtain the current training graph $\mathcal{G}_t$ and training nodes $V_t^{tr}$
    \State Select representative nodes $\mathcal{E}_t=\operatorname{DiSCo}(V_t^{tr},e)$ and construct support graph $\mathcal{G}_{sup}^{t}$
    
    \For{$j=1$ {\bfseries to} $E$}
        \State \textbf{Fast adaptation:}
        \State $\displaystyle
        \theta_{\mathrm{fast}}^{j}
        \leftarrow
        \theta^{j}
        -
        \alpha^{j}\odot
        \nabla_{\theta^{j}}
        \mathcal{L}_{sup}
        (\mathcal{G}_{sup}^{t},\theta^{j})
        $
        
        \State Compute replay loss $\mathcal{L}_{er}$ over historical subgraphs in $\mathcal{B}$
        \State Compute logit- and semantic-level distillation losses $\mathcal{L}_{logit}$ and $\mathcal{L}_{sem}$ using teacher models in $\mathcal{F}$
        \State Construct the joint meta-objective:
        \State $\displaystyle
        \mathcal{L}_{meta}
        \leftarrow
        \mathcal{L}_{\mathcal{T}_t}
        +
        \lambda_{er}\mathcal{L}_{er}
        +
        \lambda_{kd}
        \mathcal{L}_{kd}
        $
        
        \State \textbf{Meta update:}
        \State $\displaystyle
        \alpha^{j+1}
        \leftarrow
        \alpha^{j}
        -
        \eta
        \nabla_{\alpha^{j}}
        \mathcal{L}_{meta}
        (\mathcal{G}_t,\theta_{\mathrm{fast}}^{j})
        $
        \State $\displaystyle
        \theta^{j+1}
        \leftarrow
        \theta_{\mathrm{fast}}^{j}
        -
        \max(0,\alpha^{j+1})
        \odot
        \nabla_{\theta_{\mathrm{fast}}^{j}}
        \mathcal{L}_{meta}
        (\mathcal{G}_t,\theta_{\mathrm{fast}}^{j})
        $
    \EndFor
    
    \State Sampling target-type nodes and heterogeneous neighbors via $\operatorname{DiSCo}(\cdot,\cdot)$
    \State Construct the replay subgraph and update buffer $\mathcal{B}$
    \State Add the current model $f_{\theta}$ to teacher set $\mathcal{F}$ and retain the latest $K$ teachers
\EndFor

\State \textbf{Return} $f_{\theta}$
\end{algorithmic}
\end{algorithm}

\subsection{Extension to General HGNNs}\label{app:extension}

Previously, we have discussed how to align semantic-level information via knowledge distillation in meta-path-guided attention networks. In this section, we further demonstrate that the proposed method can be easily extended to HGNNs without attention mechanisms (such as RGCN), enabling the preservation of semantic information from previous tasks.

For each pair of target-type nodes $v_i, v_j \in V_{\tau_t}$, we define the attention score $e_{ij}$ based on their hidden representations from the penultimate layer of the network $h_i^{(L-1)}$ and $h_j^{(L-1)}$, using the weight matrix $W^{(L)}$ from the last layer:
\begin{equation}
    e_{ij} = \left( h_i^{(L-1)} W^{(L)} \right)^\top \tanh\left( h_j^{(L-1)} W^{(L)} \right),
\end{equation}
We then apply softmax normalization to the attention scores for each node $v_i$ to obtain an attention vector:
\begin{equation}
    \boldsymbol{\alpha}_i = \text{softmax}(e_{i1}, e_{i2}, \dots, e_{i|V_{\tau_t}|})
\end{equation}
We compute attention vectors for both the teacher model $\boldsymbol{\alpha}_i^{T}$ and the student model $\boldsymbol{\alpha}_i^{S}$, and define the semantic alignment loss as the $L_2$ distance between the two:
\begin{equation}
    \mathcal{L}_{\text{sem}} = \sum_{i \in V_{\tau_t}} \left\| \boldsymbol{\alpha}_i^{T} - \boldsymbol{\alpha}_i^{S} \right\|_2
\end{equation}

In addition to the node similarity-based attention defined above, for edge-type-aware attention models such as HGT, attention is computed for each relation type $r \in \mathcal{R}$. Let $\boldsymbol{\beta}_r^{(T)}$ and $\boldsymbol{\beta}_r^{(S)}$ denote the attention vectors for relation $r$ from the teacher and student models, respectively. The semantic alignment loss for such edge-type-based attention mechanisms is formulated as:
\begin{equation}
    \mathcal{L}_{\text{sem}} = \sum_{r \in \mathcal{R}} \left\| \boldsymbol{\beta}_r^{T} - \boldsymbol{\beta}_r^{S} \right\|_2
\end{equation}
This formulation ensures that the student model preserves relation-aware structural semantics by aligning its attention distribution with that of the teacher model on each edge type.

\section{Experiments Setup}\label{app:exp_setup}

\subsection{Datasets}\label{app:datasets}

\begin{table}[htbp]
\caption{Key characteristics  of datasets}
\label{tab:chars}
\centering
\begin{small}
\begin{tabular}{ccccccc}
\toprule
\makecell[c]{\emph{Node}\\\emph{Classification}} & \#Nodes & \makecell[c]{\#Node\\Types}
 & \#Edges & \makecell[c]{\#Edge\\Types} & \#Classes \\
\midrule
DBLP~\citep{lv2021we} & 26,128 & 4 & 239,566 & 6 & 4 \\
IMDB~\citep{lv2021we} & 21,420 & 4 & 86,642 & 6 & 5 \\
Freebase~\citep{lv2021we} & 180,098 & 8 & 1,057,688 & 36 & 7 \\
Yelp~\citep{yang2020heterogeneous} & 82,465 & 4 & 32,548,358 & 7 & 16 \\
\bottomrule
\end{tabular}
\end{small}
\end{table}
DBLP is a bibliographic dataset in the field of computer science. A widely adopted subset is used, containing four research areas, where nodes represent authors, papers, terms, and venues.

IMDB is a dataset derived from a online movie information platform. A subset including classes such as Action, Comedy, Drama, Romance, and Thriller is selected for use.

Freebase is a large-scale knowledge graph originally constructed for web search and semantic applications. A subgraph comprising eight genres of entities and approximately one million edges is sampled following procedures outlined in prior studies.

Yelp is a review network constructed from businesses, users, locations, and reviews. Although node features are not available, many business nodes are annotated with one or more labels from sixteen predefined categories. As a prototypical web application graph, Yelp embodies real-world challenges in recommendation and user–business interaction modeling.

\subsection{Baselines}\label{app:baselines}

We introduce the baseline models that we choose to compare in the following section:

\noindent \textbf{Elastic weight consolidation (EWC) \cite{kirkpatrick2017overcoming}} is a technique that regularizes the loss function, such that the model is encouraged to only modify the weights that are less important for the previous tasks. This is achieved by penalizing changes to weights that have large importance for the previously learned tasks, thereby mitigating catastrophic forgetting when learning new tasks.

\noindent \textbf{Memory aware synapses (MAS) \cite{Aljundi_2018_ECCV}} is a regularization-based method that measures the importance of parameters based on the sensitivity of predictions to these parameters. When learning a new task, the model adjusts the weights according to the network's activations to update the important parameters relevant to the new task data.

\noindent \textbf{Topology-aware weight preserving(TWP) \cite{liu2021overcoming}} is a method introduced for graph continual learning that incorporates weight preservation mechanisms based on graph topology. By considering the local structure of the graph, TWP aims to overcome catastrophic forgetting. After computing the loss, the importance scores of the model weights are calculated according to the topology of the given graph, and the loss is regularized accordingly. This method leverages the properties of the graph.

\noindent \textbf{ER-GNN \cite{zhou2021overcoming}} stores samples from previous tasks as experiences and replays them when learning new tasks. After learning task $T_i$, sample nodes are saved to the buffer using a selection function. When learning the next task $T_{i+1}$, separate graphs are constructed for each learned task $\{T_t\}_{t=1,\dots, i}$. Then, the GNN is trained using these graphs. The overall loss is calculated by regularizing the current task's loss with the loss from the separately constructed graphs through experience replay.

\noindent \textbf{MetaCLGraph \cite{unal2023meta}} combines experience replay and meta-learning. In this method, the initial parameters of the model are calculated using the current task data. When learning a new task, stored samples from previous tasks are merged with the current task data to form a new graph, which is then used to update the model parameters.

\noindent \textbf{FTF-ER \cite{pang2024ftf}} combines feature and global topological information by normalizing and weightedly integrating two types of node importance scores to evaluate node importance. Specifically, it introduces the Hodge Potential Score (HPS) module to capture global topological information. When learning a new task, a subgraph is induced using all the experience nodes stored in the buffer, the overall loss is calculated as the sum of the loss for the current task and the loss of the subgraph.

\noindent \textbf{DMSG \cite{qiao2025towards}} selects representative nodes using a diversity-aware buffer strategy that accounts for both intra-class and inter-class variations, and further augments replay through variationally generated synthetic embeddings. Adversarial and reconstruction-based objectives are used to preserve embedding quality and improve generalization during replay.

\subsection{Task Incremental HCGL}\label{app:setting}

\begin{figure}[h]
\centerline{
\includegraphics[width=0.78\linewidth]{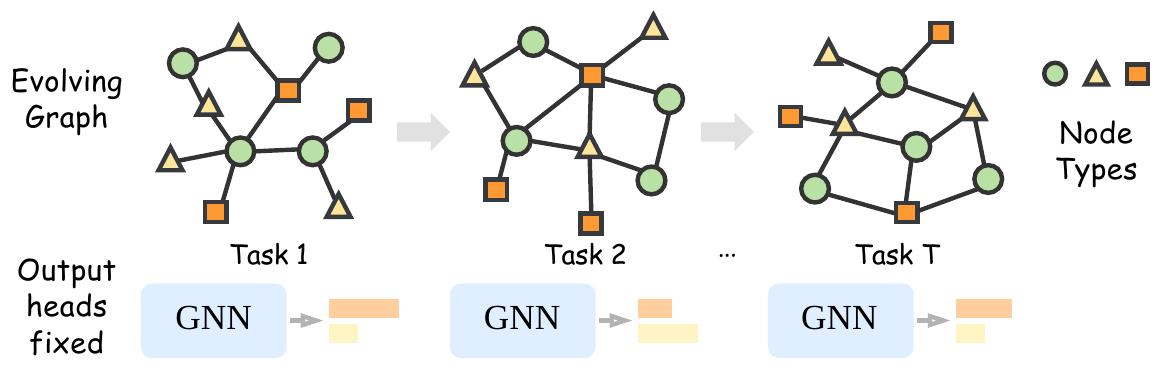}
}
\caption{Illustration of the setting used in this paper.}
\label{fig:domain_setting}
\end{figure} 

In task-IL setting, a model requires the task indicator to perform inference on a given task, and the model only needs to distinguish different classes within each task without considering the classes from existing tasks.
When the total number of classes is not divisible by $K$, the remaining class(es) are excluded from the continual task label space and are not used as target classes. Nodes belonging to the excluded target classes are removed from the constructed task sequence.
Accordingly, DBLP uses all 4 classes to form two 2-way tasks; IMDB uses 4 of its 5 classes to form two 2-way tasks; Freebase uses 6 of its 7 classes to form three 2-way tasks; and Yelp uses 15 of its 16 classes to form five 3-way tasks. For the 8-task/2-way Yelp setting, all 16 classes are used.

\subsection{Experimental Setting}\label{app:config}
Adam optimizer is used and the initial learning rate is set to 0.005 for all datasets. For HAN, each task is trained for 200 epochs with an early stopping patience of 100. For MAGNN, each task is trained for 100 epochs, with the early stopping patience set to 5 on DBLP and 10 on IMDB. For HGT, all tasks are trained for 300 epochs with a patience of 30. For datasets trained with mini-batches, the batch size is set to 8. The buffer size is uniformly set to 50 for the target node type in memory replay baselines, and 20 under mini-batch settings. For non-target node types, the buffer size is set to 200 for all memory replay methods. The hyperparameters for regularization-based baselines (e.g., EWC, MAS, and TWP) are set following the TWP official repository and the benchmark study~\citep{zhang2022cglb}. The hyper-parameters of HERO include some weight of loss $\lambda_{er},\lambda_{kd}$. We chose them from $\{0.1, 0.2, 0.3,$
$0.4, 0.5, 0.6, 0.7, 0.8, 0.9, 1.0\}$. Different combinations are adjusted for different datasets to achieve optimal performance. The hyper-parameters of the Heterogeneity-aware knowledge distillation module (HKD) include $\lambda_{logit}$ and $\lambda_{sem}$. We chose $\lambda_{logit}$ from $\{0.6, 0.7, 0.8,$
$0.9, 1.0, 1.1, 1.2, 1.3, 1.4, 1.5\}$ and set $\lambda_{sem}$ to $10.0$ globally.

\section{Scalability Analysis}\label{app:scale}

\begin{table}[t]
\centering
\small
\caption{GPU memory usage on Freebase. The first group compares
HERO with replay-based methods, while the second group provides
diagnostic controls for the additional overhead of HKD. Lower is
better.}
\label{tab:memory_scalability}
\begin{tabular}{lcc}
\toprule
Method & Avg. GPU/task (MB) $\downarrow$ & Peak GPU (MB) $\downarrow$ \\
\midrule
\multicolumn{3}{l}{\textit{Replay-based methods}} \\
\textbf{HERO (Ours)} & \textbf{7,046} & \textbf{8,107} \\
DMSG        & 7,126 & 8,131 \\
ER-GNN      & 7,065 & 8,138 \\
FTF-ER      & 7,097 & 8,122 \\
MetaCLGraph & 7,102 & 8,125 \\
\midrule
\multicolumn{3}{l}{\textit{Diagnostic controls}} \\
HERO w/o HKD & 7,013 & 8,124 \\
Finetune     & 6,973 & 8,121 \\
\bottomrule
\end{tabular}
\vspace{-2mm}
\end{table}

\begin{table}[t]
\centering
\small
\caption{Effect of the teacher-window size on Yelp. $k=T-1$
retains all historical teachers, while fixed $k$ bounds teacher
storage to $O(k)$.}
\label{tab:teacher_window}
\begin{tabular}{lcccc}
\toprule
& \multicolumn{2}{c}{Yelp 5-task}
& \multicolumn{2}{c}{Yelp 8-task} \\
\cmidrule(lr){2-3}\cmidrule(lr){4-5}
Teachers
& AP $\uparrow$ & AF $\downarrow$
& AP $\uparrow$ & AF $\downarrow$ \\
\midrule
$k=1$   & 53.51 & 13.41 & 55.55 & 11.31 \\
$\mathbf{k=3}$   & $\mathbf{56.83}$ & $\mathbf{9.29}$
        & $\mathbf{57.43}$ & $\mathbf{7.98}$ \\
$k=T-1$ & 57.44 & 8.15 & 57.81 & 7.43 \\
\bottomrule
\end{tabular}
\vspace{-2mm}
\end{table}

\paragraph{Memory and Long-Sequence Scalability.}
We further examine the scalability of HERO from both empirical memory usage and long-sequence model storage. We first measure GPU memory consumption on Freebase and compare HERO with representative replay-based methods. As shown in Table~\ref{tab:memory_scalability}, HERO requires 7,046 MB of GPU memory per task on average and reaches a peak of 8,107 MB, both slightly lower than the compared replay baselines. We further remove HKD to isolate the overhead introduced by teacher-based distillation. The average memory decreases from 7,046 MB to 7,013 MB, while Finetune requires 6,973 MB, indicating that HKD introduces only modest runtime memory overhead in the evaluated setting.

Nevertheless, retaining all historical teachers leads to a model storage cost of $O(T)$ as the number of tasks $T$ grows, even when the size of each teacher is constant. To avoid this linear growth, we maintain only the most recent $k$ teachers using a sliding window, reducing teacher storage to $O(k)$. Table~\ref{tab:teacher_window} shows the results on Yelp under 5-task and 8-task settings. Keeping three teachers provides a favorable efficiency--performance trade-off: relative to retaining all teachers, \(k=3\) reduces AP by only 0.61 and 0.38 points in the 5-task and 8-task settings, while AF increases by 1.14 and 0.55 points. In exchange, teacher storage changes from \(O(T)\) to \(O(1)\) for fixed \(k\). These results suggest that a small fixed teacher window can retain most of the benefit of multi-teacher distillation while keeping the additional model storage bounded.
Overall, HERO incurs little additional runtime memory in the evaluated setting, while the fixed teacher window prevents its teacher-storage cost from increasing with the number of tasks.

\noindent \paragraph{Training Time.} Based on the HAN backbone, we evaluate recent continual graph learning methods and the proposed HERO framework on four datasets using a single RTX 3080Ti GPU. Table~\ref{tab:train_time} reports the average training time per task. Overall, regularization-based methods such as TWP are more efficient than replay-based approaches, highlighting a known efficiency gap in replay-driven continual learning. Addressing this gap in heterogeneous settings remains an open problem. Despite relying on replay, HERO is consistently faster than other replay-based methods, showing clear gains in training efficiency. Although TWP is slightly more efficient in terms of runtime, it underperforms HERO across all benchmarks with respect to accuracy and forgetting. These measurements show manageable overhead in the evaluated settings.

\begin{table}[htbp]
\caption{Training time (s) comparison on four datasets. Lower is better. Numbers in parentheses denote the ratio relative to HERO.}

\centering
\scalebox{0.85}{
\begin{tabular}{l c c c c}
\toprule
\textbf{Method} & DBLP & IMDB & Freebase & Yelp \\
\midrule
TWP         & 0.157 (x0.42) & 0.032 (x0.78) & 0.068 (x0.59) & 3.020 (x0.50) \\
FTF-ER      & 1.619 (x4.39) & 0.238 (x5.78) & 0.351 (x3.06) & 8.424 (x1.39) \\
ER-GNN      & 0.411 (x1.11) & 0.053 (x1.29) & 0.078 (x0.68) & 6.652 (x1.10) \\
MetaCLGraph & 1.770 (x4.80) & 0.310 (x7.53) & 0.614 (x5.34) & 9.752 (x1.61) \\
HERO        & 0.369 (x1.00) & 0.041 (x1.00) & 0.115 (x1.00) & 6.043 (x1.00) \\
\bottomrule
\end{tabular}}
\vspace{-2mm}
\label{tab:train_time}
\end{table}



\section{Additional Results}\label{app:add_res}

\subsection{Results on Different Backbones}\label{app:backbone}
Additional experiment results on other two backbones: HGT~\citep{hu2020heterogeneous} and MAGNN~\citep{fu2020magnn} as shown in Table~\ref{tab:hgt_result} and \ref{tab:magnn_result}.
\begin{table}[htbp]
\centering
\small
\caption{Performance comparison on three benchmark datasets using HGT as the backbone.
The symbol $\uparrow$~($\downarrow$) indicates that higher~(lower) is better.
The best results are highlighted in \textbf{bold}, while the second-best results are \underline{underlined}.}
\setlength{\tabcolsep}{1.5mm}
\resizebox{0.9\linewidth}{!}{%
\begin{tabular}{l|cc|cc|cc}
\toprule
\multirow{2}{*}{Methods}
& \multicolumn{2}{c|}{DBLP}
& \multicolumn{2}{c|}{IMDB}
& \multicolumn{2}{c}{Freebase} \\
\cmidrule(lr){2-3}
\cmidrule(lr){4-5}
\cmidrule(lr){6-7}
& AP / \% $\uparrow$ & AF / \% $\downarrow$
& AP / \% $\uparrow$ & AF / \% $\downarrow$
& AP / \% $\uparrow$ & AF / \% $\downarrow$ \\
\midrule
Finetune
& 86.2 $\pm$ 1.5 & 12.7 $\pm$ 2.2
& 75.1 $\pm$ 7.5 & 17.0 $\pm$ 9.4
& 67.8 $\pm$ 1.8 & 16.5 $\pm$ 2.5 \\
Joint Train
& 93.9 $\pm$ 0.6 & 1.7 $\pm$ 1.0
& 80.0 $\pm$ 0.7 & 2.0 $\pm$ 0.9
& 72.5 $\pm$ 0.6 & 8.5 $\pm$ 1.2 \\
\midrule
EWC \citep{kirkpatrick2017overcoming}
& 89.1 $\pm$ 1.2 & 10.7 $\pm$ 1.8
& 77.3 $\pm$ 0.4 & 12.5 $\pm$ 0.2
& 69.6 $\pm$ 3.0 & 14.7 $\pm$ 2.9 \\
MAS \citep{Aljundi_2018_ECCV}
& 89.8 $\pm$ 2.5 & 9.0 $\pm$ 3.0
& 77.5 $\pm$ 1.2 & 11.8 $\pm$ 2.6
& \underline{70.4 $\pm$ 3.1} & 14.8 $\pm$ 2.0 \\
TWP \citep{liu2021overcoming}
& 89.1 $\pm$ 2.1 & 10.3 $\pm$ 1.1
& \underline{77.7 $\pm$ 0.6} & 11.1 $\pm$ 1.6
& 69.6 $\pm$ 2.0 & 15.1 $\pm$ 0.9 \\
ER-GNN \citep{zhou2021overcoming}
& \underline{90.7 $\pm$ 1.5} & \underline{3.5 $\pm$ 0.4}
& 77.3 $\pm$ 2.2 & 9.0 $\pm$ 3.6
& 69.5 $\pm$ 1.4 & 13.6 $\pm$ 1.7 \\
MetaCLGraph \citep{unal2023meta}
& 90.0 $\pm$ 0.7 & 10.9 $\pm$ 1.4
& 76.1 $\pm$ 0.3 & 12.2 $\pm$ 1.8
& 69.1 $\pm$ 2.2 & 16.3 $\pm$ 2.5 \\
FTF-ER \citep{pang2024ftf}
& 89.4 $\pm$ 2.2 & 5.5 $\pm$ 1.9
& 76.1 $\pm$ 1.4 & \underline{8.9 $\pm$ 3.5}
& 69.7 $\pm$ 2.8 & \underline{13.0 $\pm$ 2.5} \\
DMSG \citep{qiao2025towards}
& 88.2 $\pm$ 1.9 & 11.3 $\pm$ 2.4
& 76.1 $\pm$ 0.4 & 12.1 $\pm$ 1.5
& 64.6 $\pm$ 0.6 & 16.4 $\pm$ 2.7 \\
\rowcolor{gray!20}
HERO
& \textbf{92.8 $\pm$ 0.8} & \textbf{3.3 $\pm$ 1.2}
& \textbf{78.4 $\pm$ 0.4} & \textbf{6.6 $\pm$ 1.7}
& \textbf{70.6 $\pm$ 1.2} & \textbf{9.1 $\pm$ 2.7} \\
\bottomrule
\end{tabular}}
\vspace{-2mm}
\label{tab:hgt_result}
\end{table}

\begin{table}[htbp]
\centering
\small
\caption{Performance comparison on DBLP and IMDB using MAGNN as the backbone.
MAGNN runs out of memory on Freebase and is therefore not reported.
The symbol $\uparrow$~($\downarrow$) indicates that higher~(lower) is better.
The best results are highlighted in \textbf{bold}, while the second-best results are \underline{underlined}.}
\setlength{\tabcolsep}{2mm}
\resizebox{0.73\linewidth}{!}{%
\begin{tabular}{l|cc|cc}
\toprule
\multirow{2}{*}{Methods}
& \multicolumn{2}{c|}{DBLP}
& \multicolumn{2}{c}{IMDB} \\
\cmidrule(lr){2-3}
\cmidrule(lr){4-5}
& AP / \% $\uparrow$ & AF / \% $\downarrow$
& AP / \% $\uparrow$ & AF / \% $\downarrow$ \\
\midrule
Finetune
& 79.1 $\pm$ 5.1 & 34.8 $\pm$ 10.1
& 71.3 $\pm$ 0.6 & 21.7 $\pm$ 3.0 \\
Joint Train
& 94.1 $\pm$ 0.1 & 1.9 $\pm$ 1.0
& 77.4 $\pm$ 0.7 & 4.7 $\pm$ 1.5 \\
\midrule
EWC \citep{kirkpatrick2017overcoming}
& 91.0 $\pm$ 2.3 & 10.9 $\pm$ 4.5
& 73.4 $\pm$ 1.6 & 16.7 $\pm$ 2.5 \\
MAS \citep{Aljundi_2018_ECCV}
& 92.1 $\pm$ 2.1 & 7.8 $\pm$ 5.3
& 74.3 $\pm$ 2.0 & 14.2 $\pm$ 3.7 \\
TWP \citep{liu2021overcoming}
& 91.6 $\pm$ 2.1 & 7.4 $\pm$ 5.0
& 74.0 $\pm$ 0.8 & 13.7 $\pm$ 3.3 \\
ER-GNN \citep{zhou2021overcoming}
& 90.0 $\pm$ 2.1 & 12.4 $\pm$ 4.3
& 75.3 $\pm$ 0.9 & 10.3 $\pm$ 1.5 \\
MetaCLGraph \citep{unal2023meta}
& 92.3 $\pm$ 0.4 & \underline{2.1 $\pm$ 0.7}
& \underline{75.6 $\pm$ 1.0} & 11.8 $\pm$ 1.3 \\
FTF-ER \citep{pang2024ftf}
& \underline{92.8 $\pm$ 0.8} & \textbf{1.9 $\pm$ 2.3}
& 73.0 $\pm$ 0.6 & 15.8 $\pm$ 1.4 \\
DMSG \citep{qiao2025towards}
& 90.6 $\pm$ 1.1 & 8.6 $\pm$ 2.2
& 74.5 $\pm$ 0.5 & \underline{8.8 $\pm$ 2.7} \\
\rowcolor{gray!20}
HERO
& \textbf{93.2 $\pm$ 0.5} & 3.1 $\pm$ 0.8
& \textbf{76.8 $\pm$ 1.1} & \textbf{7.5 $\pm$ 1.4} \\

\bottomrule
\end{tabular}}
\vspace{-2mm}
\label{tab:magnn_result}
\end{table}

\subsection{Reordered Task Sequence}
\label{app:exp_robust}
We further evaluate robustness under task-order variations by training with randomly permuted task sequences on DBLP, IMDB, and Freebase. Results is reported in Table~\ref{tab:reorder_seq}. 
Under the tested random task orders, HERO improves AP over the strongest baseline by 0.9 points on DBLP, 2.1 points on IMDB, and 5.1 points on Freebase. HERO also obtains the lowest AF on IMDB and Freebase, while FTF-ER gives a lower AF on DBLP. These results show that HERO's AP advantage persists under the evaluated task permutations.

\begin{table}[htbp]
\centering
\small
\caption{Performance comparison under reordered task sequences on three benchmark datasets using HAN as the backbone.
The symbol $\uparrow$~($\downarrow$) indicates that higher~(lower) is better.
The best results are highlighted in \textbf{bold}, while the second-best results are \underline{underlined}.}
\label{tab:reorder_seq}
\setlength{\tabcolsep}{1.5mm}
\resizebox{0.95\linewidth}{!}{%
\begin{tabular}{l|cc|cc|cc}
\toprule
\multirow{2}{*}{Methods}
& \multicolumn{2}{c|}{DBLP}
& \multicolumn{2}{c|}{IMDB}
& \multicolumn{2}{c}{Freebase} \\
\cmidrule(lr){2-3}
\cmidrule(lr){4-5}
\cmidrule(lr){6-7}
& AP / \% $\uparrow$ & AF / \% $\downarrow$
& AP / \% $\uparrow$ & AF / \% $\downarrow$
& AP / \% $\uparrow$ & AF / \% $\downarrow$ \\
\midrule
Finetune
& 83.5 $\pm$ 0.9 & 24.3 $\pm$ 2.1
& 56.7 $\pm$ 0.7 & 49.8 $\pm$ 1.3
& 49.2 $\pm$ 2.3 & 31.8 $\pm$ 5.7 \\
EWC \citep{kirkpatrick2017overcoming}
& 90.7 $\pm$ 1.1 & 10.5 $\pm$ 2.5
& 67.8 $\pm$ 2.5 & 24.0 $\pm$ 2.6
& 54.0 $\pm$ 11.0 & 24.2 $\pm$ 20.1 \\
MAS \citep{Aljundi_2018_ECCV}
& 92.6 $\pm$ 2.5 & 4.8 $\pm$ 5.3
& 72.0 $\pm$ 4.9 & 13.0 $\pm$ 10.7
& 47.1 $\pm$ 0.6 & 34.6 $\pm$ 2.3 \\
TWP \citep{liu2021overcoming}
& 93.6 $\pm$ 1.1 & 3.5 $\pm$ 1.5
& \underline{73.4 $\pm$ 0.5} & 11.2 $\pm$ 4.9
& 52.0 $\pm$ 6.8 & 26.4 $\pm$ 9.9 \\
ER-GNN \citep{zhou2021overcoming}
& 93.2 $\pm$ 2.7 & 6.1 $\pm$ 5.4
& 69.7 $\pm$ 2.7 & 13.6 $\pm$ 3.2
& \underline{56.0 $\pm$ 3.9} & \underline{17.7 $\pm$ 7.5} \\
FTF-ER \citep{pang2024ftf}
& 94.6 $\pm$ 0.5 & \textbf{1.1 $\pm$ 0.5}
& 67.2 $\pm$ 4.7 & 30.4 $\pm$ 3.9
& 45.1 $\pm$ 1.1 & 35.6 $\pm$ 0.7 \\
MetaCLGraph \citep{unal2023meta}
& \underline{94.7} $\pm$ 0.5 & 3.4 $\pm$ 0.8
& 65.6 $\pm$ 3.6 & 26.9 $\pm$ 4.8
& 50.1 $\pm$ 2.1 & 26.2 $\pm$ 4.5 \\
\rowcolor{gray!20}
HERO
& \textbf{95.6 $\pm$ 0.6} & \underline{2.1 $\pm$ 2.6}
& \textbf{75.5 $\pm$ 0.1} & \textbf{8.0 $\pm$ 0.3}
& \textbf{61.1 $\pm$ 1.0} & \textbf{5.1 $\pm$ 1.5} \\

\bottomrule
\end{tabular}}
\vspace{-2mm}
\end{table}

\subsection{Further Ablation Study on HKD}\label{app:hkd_abl}
\begin{figure}
\centering
\includegraphics[width=0.75\linewidth]{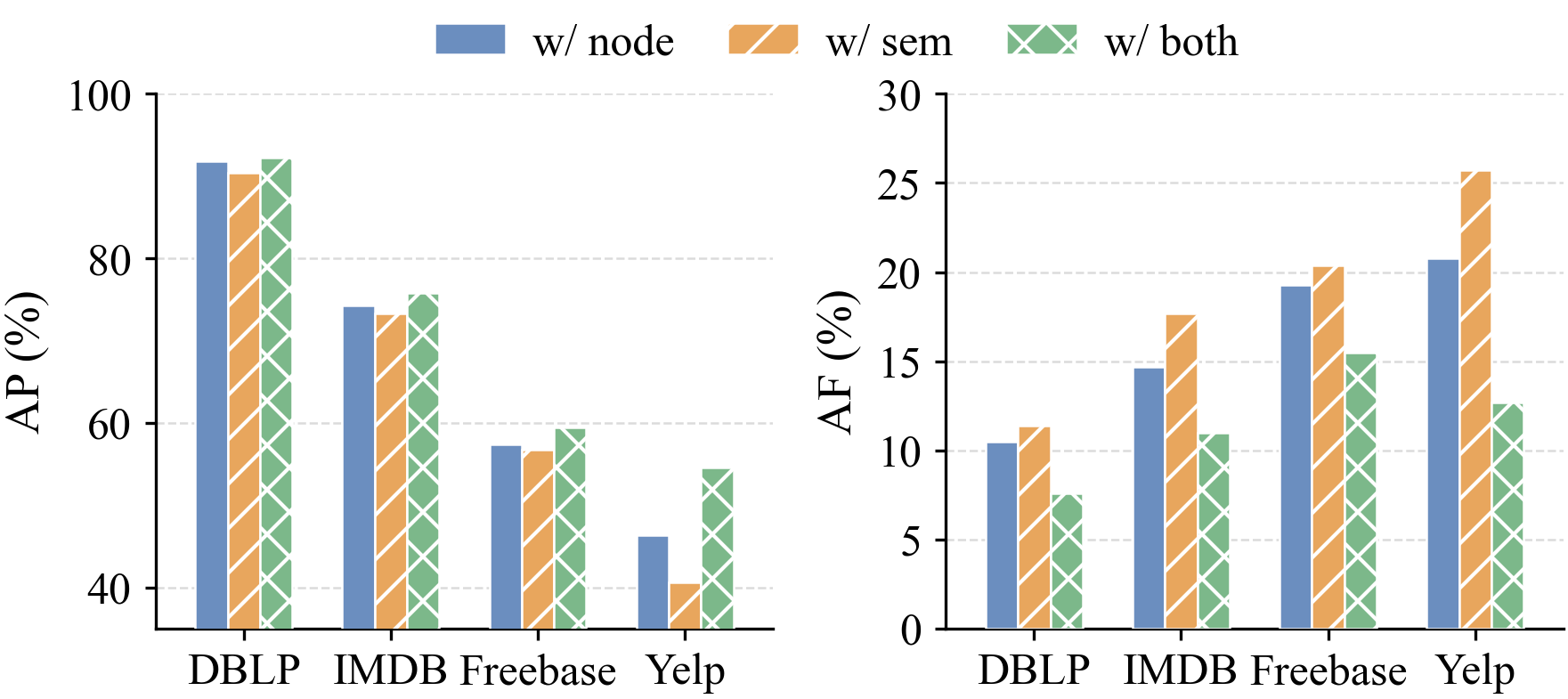}

\caption{Ablation Study on Heterogeneity-aware Knowledge Distillation Module on all datasets.}

\label{fig:hkd_ablation}
\end{figure}

We further analyze the contribution of the two submodules in the HKD module: logit-level distillation and semantic-level distillation to overall model performance, and assess the effectiveness of our proposed design.
Figure~\ref{fig:hkd_ablation} illustrates the impact of different distillation strategies—node-level only (“with node”), semantic-level only (“with sem”), and the combination of both (“with both”)—on the model’s Average Performance (AP) and Average Forgetting (AF) across four datasets (DBLP, IMDB, Freebase, Yelp). The results show that the combined strategy (“with both”) consistently achieves the highest AP and lowest AF across all datasets, demonstrating that logit-level and semantic-level distillation complement each other. Semantic-level distillation alone (“with sem”) exhibits worse forgetting suppression on Freebase and Yelp, suggesting it may struggle to independently capture task-level semantic changes. On complex heterogeneous graphs such as Yelp, the synergy of both distillation levels yields especially significant performance improvements, indicating that addressing both structural and semantic forgetting is crucial.

\subsection{Forgetting-aware Gap}
\label{app:fag}

AP (Average Performance) measures the mean test accuracy across all previously learned tasks, while AF (Average Forgetting) quantifies the performance drop on a specific task after learning subsequent ones. However, we observe that in order to retain knowledge from earlier tasks, the model often compromises its performance on the final task. To assess this trade-off, we introduce a new metric, \textbf{Forgetting-aware Gap} (FaG), defined as the difference between the test accuracy obtained by training the model solely on the final task and the test accuracy on the same task after completing all tasks under the continual learning framework:
\begin{equation}
    \text{FaG} = a_T^{FT} - a_T^{CL}
\end{equation}
where $a_T$ denotes the test accuracy on the final task $\mathcal{T}_T$, with “FT” referring to the finetuning baseline (i.e., training only on task $\mathcal{T}_T$), and “CL” referring to the accuracy obtained after training with a continual learning method.
This gap reflects the performance degradation caused by the inherent plasticity-stability trade-off in continual learning. Due to limited space, we provide experiment results in Appendix~\ref{app:fag}.

\begin{table}[htbp]
\centering
\small
\caption{Forgetting-aware Gap (FaG, \%) on four benchmark datasets, averaged over five runs. 
Lower values indicate better performance. 
The best results are highlighted in \textbf{bold}, while the second-best results are \underline{underlined}.}
\label{tab:fag}
\setlength{\tabcolsep}{4mm}
\begin{tabular}{lcccc}
\toprule
\textbf{Methods} 
& \textbf{DBLP} 
& \textbf{IMDB} 
& \textbf{Freebase} 
& \textbf{Yelp} \\
\midrule

\multicolumn{5}{l}{\textit{Other}} \\
Joint Train 
& 0.69 & 2.47 & 11.59 & 11.58 \\
TWP 
& 1.03 & \underline{2.27} & 10.44 & \textbf{5.58} \\

\midrule
\multicolumn{5}{l}{\textit{Memory Replay}} \\
ER-GNN 
& \textbf{0.22} & 2.62 & \textbf{8.44} & 8.46 \\
FTF-ER 
& 0.83 & 3.43 & 10.59 & 7.19 \\
MetaCLGraph 
& 1.36 & 2.87 & 13.68 & 13.61 \\
DMSG
& 0.89 & 3.76 & 11.87 & 15.57 \\
\rowcolor{gray!20}
HERO 
& \underline{0.63} & \textbf{1.31} & \underline{9.90} & \underline{6.88} \\

\bottomrule
\end{tabular}
\vspace{-2mm}
\end{table}

As shown in Table~\ref{tab:fag}, HERO achieves a consistently competitive forgetting-aware gap across different datasets. It obtains the lowest FaG on IMDB and the second-lowest results on DBLP, Freebase, and Yelp. These results suggest that HERO maintains a favorable balance between retaining previously learned knowledge and adapting to new tasks, rather than improving average performance at the cost of excessive forgetting.

\subsection{Additional Parameter Analysis and Visualizations}\label{app:vis}
\begin{figure}[htbp]
\centering
\vspace{-2mm}
\includegraphics[width=0.98\linewidth]{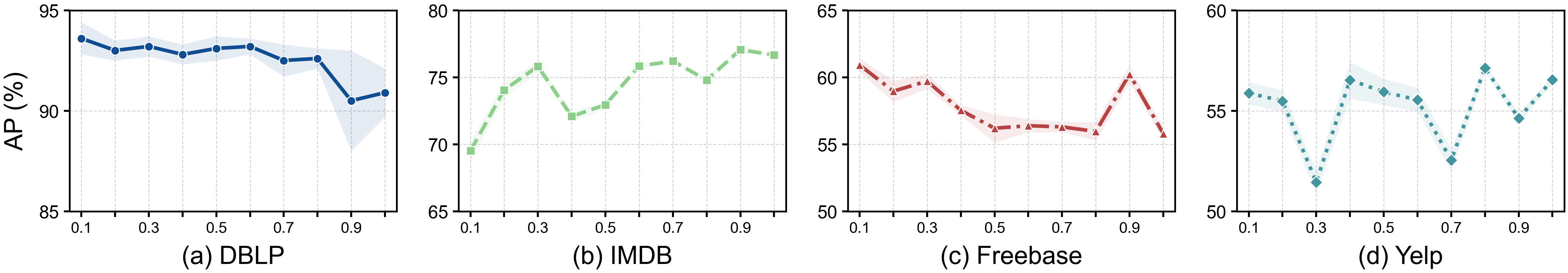}\vspace{-2mm}
\caption{Effect of the distillation loss weight $\lambda_{kd}$ across four datasets.} 
\vspace{-2mm}
\label{fig:lkd}
\end{figure}

\begin{figure}[htpb]
\centering
\vspace{-2mm}
\includegraphics[width=0.98\linewidth]{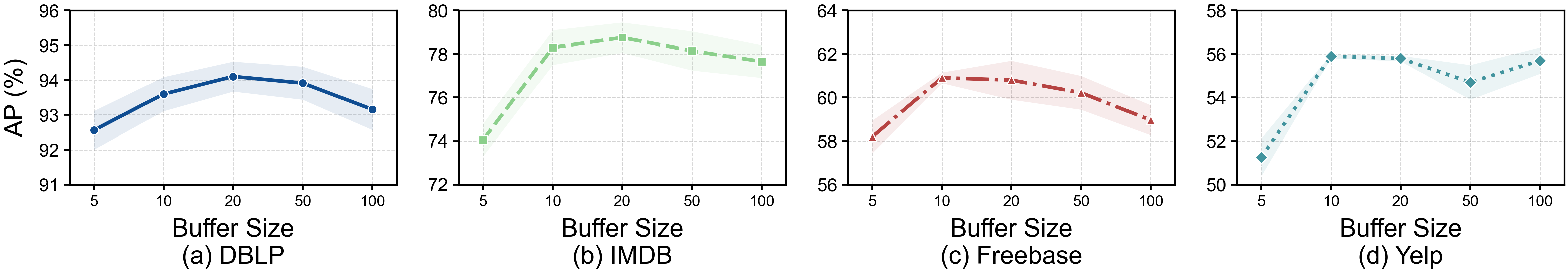}\vspace{-2mm}
\caption{Effect of the buffer size across four datasets.} 
\vspace{-2mm}
\label{fig:budget}
\end{figure}

We provide additional sensitivity analyses for two hyperparameters: the distillation loss weight $\lambda_{kd}$ and the shot number as shown in Figure~\ref{fig:lkd} and \ref{fig:budget}.

\section{Extended Related Work}
\label{app:related_work}
\vspace{-3mm}

\noindent\paragraph{\textbf{Heterogeneous Graph Neural Networks.}}
Heterogeneous Graph Neural Networks (HGNNs) are designed to model graphs containing multiple node and edge types by encoding type-dependent structures and semantics. Representative HGNNs include metapath-guided methods, which aggregate information along predefined or learned semantic paths~\citep{fu2020magnn,zhang2019heterogeneous, wang2019heterogeneous,schlichtkrull2018modeling,dong2017metapath2vec, sun2011pathsim,yun2019graph,yang2023simple,zhu2024hagnn}, and metapath-free methods, which directly model interactions between node and relation types~\citep{hu2020heterogeneous,lv2021we,
zhu2019relation,hong2020attention,zhang2022simple,fu2023multiplex}. These methods have shown strong performance in extracting heterogeneous structural and semantic information. Recent work such as G-Substrate~\cite{li2026graph} further explores reusable graph representations across different data modalities and learning contexts, highlighting the value of preserving transferable structural knowledge.

Most existing HGNN studies adopt an offline learning protocol in which the training graph is jointly accessible. Under continual learning, however, an HGNN must update its parameters using the current task and a bounded historical memory. This creates a different preservation problem because the prediction of a target node depends not only on its own feature and label, but also on typed neighbors, relation-specific transformations, and semantic aggregation channels. Our work does not introduce a new HGNN encoder. Instead, it studies how representative HGNN backbones can retain these heterogeneous computation patterns when learned over a sequence of class-disjoint tasks.

\noindent\paragraph{\textbf{Continual Graph Learning.}}
Continual Graph Learning (CGL) studies how graph models learn from sequentially presented data while reducing catastrophic forgetting~\citep{zhang2022cglb,zhang2024continual}. Existing methods can be broadly grouped into parameter regularization or isolation, experience replay, and knowledge distillation. Regularization- and parameter-isolation-based methods protect knowledge by restricting updates to parameters estimated to be important for previous tasks~\citep{zhang2022hierarchical,
zhang2023continual,kirkpatrick2017overcoming,Aljundi_2018_ECCV,
liu2021overcoming}. Some graph-specific methods further incorporate topological information when estimating parameter importance. More recently, DRIFT~\citep{sun2026drift} introduced a task-free continual graph learning benchmark that models continuous distribution shifts without assuming explicit task boundaries, extending the evaluation of continual graph learning beyond conventional task-based settings.
However, in an HGNN, the same parameter may play different roles for different node types, relation types, or semantic channels. A globally aggregated importance score does not directly represent such type- or relation-conditioned roles. It may therefore over-constrain shared parameters or provide insufficient protection for low-frequency heterogeneous patterns. Experience-replay methods retain historical nodes, embeddings, or subgraphs and revisit them when learning subsequent tasks~\citep{zhang2023ricci,ahrabian2021structure,
zhou2021overcoming,zhang2022sparsified,choi2024dslr,
zhang2024topology,hoang2023universal,han2024topology,
song2025exploring}. These methods have shown that data selection and topology preservation are important for graph continual learning. Nevertheless, replay in heterogeneous graphs introduces an additional context-reconstruction problem. A target node with the same feature and label can produce different representations when its typed neighbors or relation-specific connections change. Storing isolated target nodes may therefore fail to reproduce the heterogeneous computation that generated the historical training signal. On the other hand, retaining complete heterogeneous neighborhoods can lead to substantial memory and computation costs. DiSCo addresses this trade-off by combining representative target-node selection with bounded multi-type neighbor expansion. 
Distillation-based graph continual learning methods preserve historical knowledge by aligning model predictions or representations across learning stages~\citep{Xu2020GraphSAIL,dong2021rkd,qiu2025efficient}. Such methods provide useful functional constraints, but output or embedding alignment does not explicitly determine which relation- or metapath-specific semantic channels produce the prediction. Two HGNNs may generate similar outputs while assigning different importance to heterogeneous semantic contexts. HERO therefore combines logit-level distillation with semantic-level alignment to constrain both historical predictions and heterogeneous semantic responses. Different from settings that explicitly introduce new node or relation types, our protocol retains a fixed heterogeneous schema and partitions target classes into a sequence of disjoint tasks. The main challenge studied here is thus the preservation of type-dependent and relation-dependent computation under sequential parameter updates and bounded historical access.

\noindent\paragraph{\textbf{Meta-Learning.}}
Meta-learning aims to improve adaptation to new tasks by transferring knowledge across related learning episodes. Existing approaches are commonly divided into optimization-based methods, such as MAML~\citep{finn2017model} and Reptile~\citep{nichol2018reptile}, and metric-based methods, such as Prototypical Networks~\citep{snell2017prototypical} and Matching Networks~\citep{vinyals2016matching}. Optimization-based methods typically learn an initialization or update mechanism from which a model can adapt using a small number of gradient steps.
Meta-learning has also been studied in continual learning. La-MAML~\citep{gupta2020look} uses look-ahead optimization to account for the influence of current updates on historical examples, while representation-based approaches learn features that remain adaptable across sequential tasks~\citep{javed2019meta}. In graph learning, MetaCLGraph~\citep{unal2023meta} combines meta-learning with experience replay for continual graph learning, whereas HG-Meta~\citep{zhang2022hg} studies few-shot adaptation on heterogeneous graphs.
HERO follows the look-ahead optimization principle but focuses on heterogeneous continual node classification. Its adaptation module uses a small representative support graph to obtain a fast update, and evaluates this update using a joint objective containing the current-task, replay, and distillation constraints. Thus, meta-adaptation primarily improves plasticity on the incoming task, while DiSCo and heterogeneity-aware distillation are responsible for historical knowledge retention. 

\end{document}